%% file: main_arxiv_5304.tex
\documentclass[10pt,twocolumn,letterpaper]{article}

\usepackage{cvpr}
\usepackage{times}
\usepackage{epsfig}
\usepackage{graphicx}
\usepackage{amsmath}
\usepackage{amssymb}
\usepackage[utf8]{inputenc} 
\usepackage[T1]{fontenc}    
\usepackage{url}            
\usepackage{booktabs}       
\usepackage{amsfonts}       
\usepackage{nicefrac}       
\usepackage{microtype}      
\usepackage{textcomp}      	
\usepackage{wrapfig}
\usepackage{multirow}
\usepackage{xcolor}
\usepackage[export]{adjustbox}

\include{def2}

\definecolor{1st}{rgb}{0.1, 0.1, 1.0}
\definecolor{2nd}{rgb}{0.0, 0.6, 1.0}

\newcommand{\rbold}[1]{\textbf{\textcolor{1st}{#1}}}
\newcommand{\bbold}[1]{\textbf{\textcolor{2nd}{#1}}}

\usepackage[pagebackref=true,breaklinks=true,letterpaper=true,colorlinks,bookmarks=false]{hyperref}

\cvprfinalcopy 

\def\cvprPaperID{5304} 

\ifcvprfinal\fi
 
\begin{document}
 
\title{Variational Prototyping-Encoder: One-Shot Learning with Prototypical Images}

%

\author{
  Junsik Kim\qquad Tae-Hyun Oh${}^\dagger$
  \qquad Seokju Lee\qquad Fei Pan\qquad In So Kweon\\
Dept. of Electrical Engineering, KAIST, Daejeon, Korea\\
${}^\dagger$MIT CSAIL, Cambridge, US\\
}


\maketitle
\thispagestyle{empty}
\begin{abstract}

In daily life, graphic symbols, such as traffic signs and brand logos, are 
ubiquitously utilized around us due to its intuitive expression beyond language boundary.
We tackle an open-set graphic symbol recognition problem by one-shot classification with prototypical images as a single training example for each novel class.
We take an approach to learn a generalizable embedding space for novel tasks. 
We propose a new approach called variational prototyping-encoder (VPE) that learns the image translation task from real-world input images to their corresponding prototypical images as a meta-task.
As a result, VPE learns image similarity as well as prototypical concepts which differs from widely used metric learning based approaches.
Our experiments with diverse datasets demonstrate that the proposed VPE performs favorably against competing metric learning based one-shot methods.
Also, our qualitative analyses show that our meta-task induces an effective embedding space suitable for unseen data representation.

\end{abstract}

\input{01_Introduction}
\input{02_Related_work}

\input{03_Proposed_method}

\input{04_Experiment}

\input{05_Conclusion}

{\flushleft \bf Acknowledgement } This work was supported by the Technology Innovation Program (No. 10048320), funded by the Ministry of Trade, Industry \& Energy (MI, Korea).

{\small
\bibliographystyle{ieee}
\bibliography{egbib}
}


\cleardoublepage

\onecolumn
\input{06_Supple}

\end{document}

%% file: def2.tex
\newcommand{\bt}{{\mathbf{t}}}

\newcommand{\bx}{{\mathbf{x}}}

\newcommand{\bz}{{\mathbf{z}}}

\newcommand{\bI}{\mathbf{I}}

\newcommand{\calL}{{\mathcal{L}}}

\newcommand{\calN}{{\mathcal{N}}}

\newcommand{\calX}{{\mathcal{X}}}

\newcommand{\bepsilon}{\mbox{\boldmath $\epsilon$}}

\newcommand{\bmu}{\mbox{\boldmath $\mu$}}

\newcommand{\bsigma}{\mbox{\boldmath $\sigma$}}

\newcommand{\be}{\begin{eqnarray}}
\newcommand{\ee}{\end{eqnarray}}
\newcommand{\bee}{\begin{eqnarray*}}
\newcommand{\eee}{\end{eqnarray*}}

\newcommand{\matrixb}{\left[ \begin{array}}
\newcommand{\matrixe}{\end{array} \right]}

\newcommand{\Tref}[1]{Table~\ref{#1}}
\newcommand{\Eref}[1]{Eq.~(\ref{#1})}
\newcommand{\Fref}[1]{Fig.~\ref{#1}}

\newcommand{\Cref}[1]{Chap.~\ref{#1}}

\def\eg{\emph{e.g.}}

\def\etal{\emph{et al.}}
\def\ie{\emph{i.e.}}

%% file: 01_Introduction.tex
\section{Introduction}
\label{sec:intro}


A meaningful graphic symbol visually and compactly expresses semantic information. 
Such graphic symbols are called ideogram,\footnote{This is also formally called as a pictogram, pictogramme, pictograph, simply picto or icon. In this work, we interchangeably refer to an ideogram using the word ``symbol'' for simplicity.}
which are designed to encode signal or identity information in an abstract form.
They effectively convey the gist of intended signals while capturing the attention of the reader in a way that allows the reader to grasp the ideas readily and rapidly~\cite{borkin2016beyond}.
Its instant (immediate) recognition characteristic is leveraged for safety signals (e.g., traffic signs) and for better visibility and identity of commercial logos.
Moreover, the compactness of iconic representativeness enables emoticons and visual hashtags~\cite{bylinskii2017understanding}.
Ideograms are often independent of any particular language and are comprehensible only by those with familiarity with prior conventions beyond language boundaries, \eg, pictorial resemblance to a physical object.



\begin{figure}
\centering
 {\includegraphics[width=1.0\linewidth]{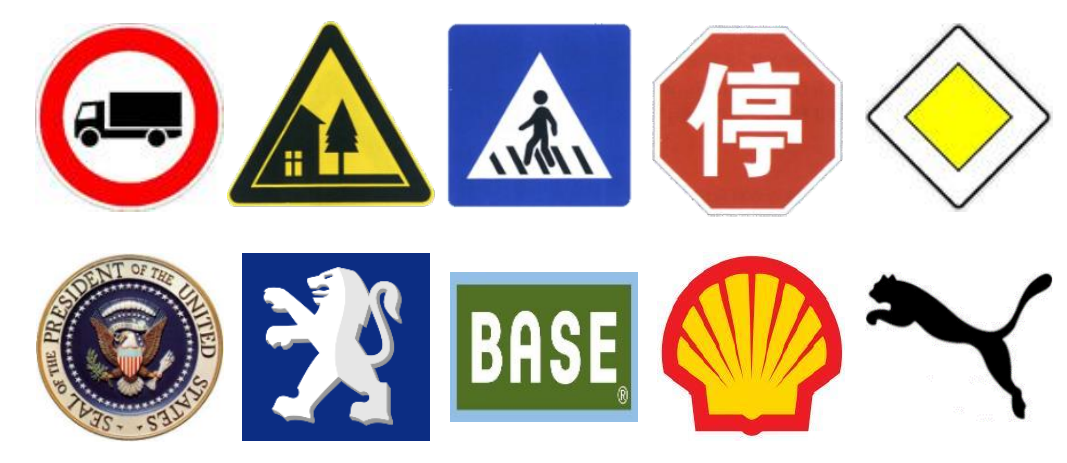}}
 
  \caption{Prototypes of symbolic icons. The top and bottom rows show traffic signs and logo prototypes, respectively.}
  \vspace{-5mm}
  \label{fig:logos}
\end{figure}

While such symbols utilize human-perception-friendly designs, machine-based understanding of the abstract visual imagery is not necessarily straightforward due to several challenges.
Original symbols in a canonical domain as shown in \Fref{fig:logos}, referred to as a prototype, are rendered in a physical form by printing or displaying.
These prototypes go through geometric and photometric perturbations via printing and imaging pipelines.
The discrepancy between real and canonical domains introduces a large perceptual gap in the visual domain (termed \emph{domain discrepancy}).
This gap is significant in that it is difficult to close it due to extreme data imbalance between real images and a single prototype of a symbol (called an \emph{intra-class data imbalance}).
Moreover, even for real images, the annotation is typically expensive when constructing a large-scale real dataset.
Although there are a few datasets with a limited number of classes, they have a noticeable class imbalance (called an \emph{inter-class data imbalance}).
Thereby, the absence of a large number of training examples for a class often raises an issue when training a large capacity learner, \ie, deep neural networks.

\begin{figure*}[t]
	\centering	
{\includegraphics[width=1.0\linewidth]{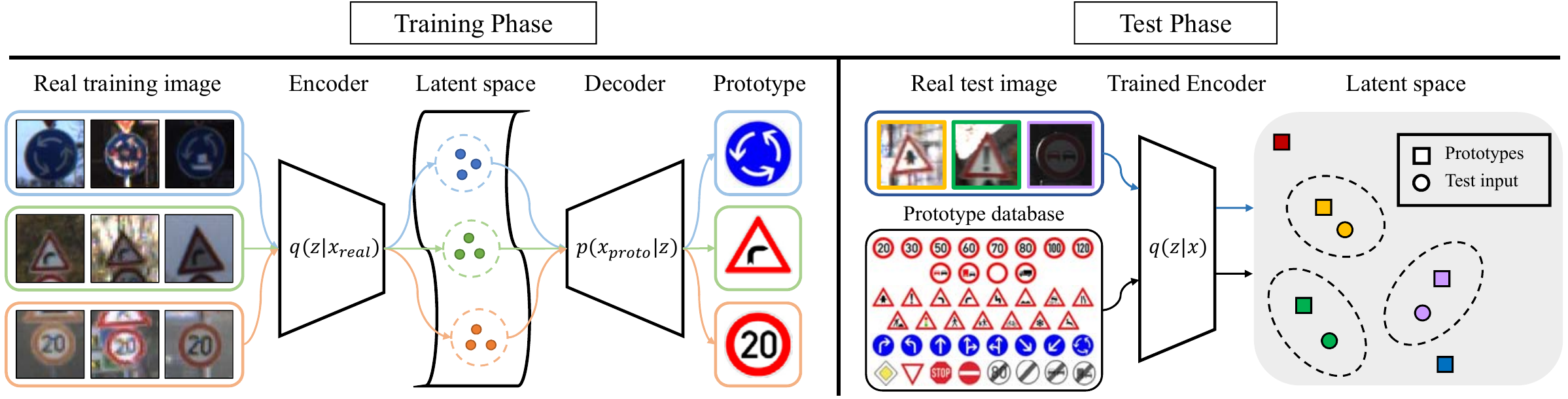}}\hspace{2mm}
	\vspace{-3mm}
	\caption{Illustration of the training and test phases of the variational prototyping-encoder. During training, the encoder encodes real domain input images to latent distribution $q(z|x)$. The decoder then reconstructs the encoded distribution back to a prototype that corresponds to the input image. In the test phase, the trained encoder is used as a feature extractor. Test images and prototypes in the database are encoded into the latent space. We then perform nearest neighbor classification to classify the test images. Note that classes of the prototypes in the test phase database are not used in the training phase, \ie, novel classes.}
	\label{fig:VAE_oneshot}
	\vspace{-3mm}
\end{figure*}

To deal with such challenges, in this work we present a deep neural network called variational prototyping-encoder (VPE) for one-shot classification of graphic symbols.
Given a single prototype of each symbol class (called a support set), VPE classifies a query into its corresponding category without requiring a large fully supervised dataset, \ie, one-shot classification.
The key ideas when attempting to alleviate the domain discrepancy and data imbalance issues are as follows: 1) VPE exploits existing pairs of prototypes and their corresponding real images to learn a  generalizable latent space for unseen class data. 
2) Instead of introducing a pre-determined metric, VPE learns an image translation~\cite{isola2017image} but from real images to prototype images, whereby the prototype is used as a strong supervision signal with high level visual appearance knowledge. 3) VPE leverages a variational autoencoder (VAE)~\cite{kingma2013auto} structure to induce a latent feature space implicitly, where the features from real data form a compact cluster around a feature point of the corresponding prototype.
This is illustrated in \Fref{fig:VAE_oneshot}.

In the test phase, as was typically done in prior works \cite{koch2015siamese,kim2018co,vinyals2016matching,snell2017prototypical},
we can easily classify queries by means of a simple nearest neighbor (NN) classification scheme in the learned latent space, where the distances between a real image feature and the given prototype features are measured and the class closest to the input feature is assigned.
For test purposes, we evaluate the prototypes from unseen categories in the test phase.
Our method can also be used for open set classification, as an unlimited number of prototypical classes can be dealt with by regarding prototypes as an open set database.

Through empirical experimental assessments of various one-shot evaluation scenarios, we show that the proposed model performs favorably against recent metric-based one-shot learners. The improvement on traffic sign datasets is noticeably significant compared to the second best method (53.30\%$\to$83.79\% on the GTSRB scenario and 58.75\%$\to$71.80\% on the GTSRB$\to$TT100K scenario) as well as on logo datasets (40.95\%$\to$53.53\% on the Belga$\to$Flickr32 scenario and 36.62\%$\to$57.75\% on the  Belga$\to$Toplogos scenario).
We also provide a visual understanding of VPE's embedding space by plotting t-SNE feature distributions and the average images of top-K retrieved images. The source code is publicly available. \footnote{\url{https://github.com/mibastro/VPE}}

%% file: 02_Related_work.tex
\section{Related Work}
In the one-shot learning context, the pioneering works of Fei-Fei~\etal~\cite{fe2003bayesian} 
hypothesize that efficiency of learning in humans may come from the advantage of prior experience. To mimic this property, they explored a Bayesian framework to learn generic \emph{prior} knowledge from unrelated tasks, which can be quickly adapted to new tasks with few examples and forms the \emph{posterior}.
More recently, Lake~\etal~\cite{lake2015human} developed a method of learning the concepts of the generative process with simple examples by means of hierarchical Bayesian program learning, where the learned concepts are also readily generalizable to novel cases, even with a single example.
Despite the success of recent end-to-end deep neural networks (DNN) in other learning tasks, one-shot learning remains a persistently challenging problem, and hand-designed systems often outperform DNN based methods~\cite{lake2015human}.

Nonetheless, in one-shot learning (including few-shot learning), the efforts to exploit the benefit of DNN is under progression.
One-shot learning regime is inherently harsh due to the over-fitting issue caused by a low number of data.
Thus, recent DNN based approaches have mainly been progressed either to achieve generalizable metric space with regard to unrelated task data (\ie, embedding space learning) or to learn high-level strategies (\ie, meta-learning).

Our method is close to the former category. 
Once a metric is given, non-parametric models such as the nearest neighbor (NN) enable unseen examples to be assimilated instantly without re-training; hence, novel category classification can be done by a simple NN.
The following works are related: metric learning by Siamese networks~\cite{koch2015siamese}, Quadruplet networks~\cite{kim2018co} and N-way metric learning~\cite{vinyals2016matching,snell2017prototypical}.
Given a metric (\eg, Euclidean distance~\cite{koch2015siamese,kim2018co,snell2017prototypical}, cosine distance~\cite{vinyals2016matching}), these approaches learn an embedding space (latent space) in the hope of generalization to novel but related domain data.
Our method is different in that we do not specify a metric directly but implicitly learn an embedding space by a meta-task, \ie, image translation from a real domain image to a prototype image.

Recent meta-learning approaches have been applied to few-shot learning.
Santoro~\etal~\cite{santoro2016meta} and Mishra~\etal~\cite{mishra2017simple} take a sequence learning approach as a meta-learner so that given a series of input sequences, the learner learns high-level strategies by which possibly to solve new tasks.
Ravi \& Larochelle~\cite{ravi2017optimization} and Finn \etal~\cite{finn2017model} seek to learn a representation that can be easily fine-tuned to new data with a few steps of gradient descent updates.
Given that most meta-learner based methods~\cite{santoro2016meta,mishra2017simple,ravi2017optimization,finn2017model,vinyals2016matching,snell2017prototypical} learn high-level strategies, they typically adopt episodic training schemes that must be well-coordinated.
This is contrary to the aforementioned metric learning based approaches~\cite{koch2015siamese,kim2018co} including our method, where the training steps are usually rather straightforward.

The methods discussed above focus on cases in which examples in supported set and a query are from the same domain.
In our problem setup, the significant discrepancy between real-world query images and the prototype in the support set introduces new challenges.
There have been few attempts related to one-shot learning with prototypes. Jetley~\etal~\cite{jetley2015prototypical} proposed a feature transform approach to align features of real images with pre-defined hand-crafted features of prototypes. 
Kim~\etal~\cite{kim2018co} is the work closest to our method. They proposed the learning of co-domain embedding using deep quadruplet networks in an end-to-end manner so that an embedding of prototype and real-world images are mapped into a common feature space.  
Recently, Snell~\etal~\cite{snell2017prototypical} proposed prototypical networks for few-shot learning in an extension of Vinyals~\etal~\cite{vinyals2016matching}. However, their definition of a prototype differs from ours in that their prototype is defined according to the mean centroid of a class on the same domain with queries, while our prototype is a prototypical image.

%% file: 03_Proposed_method.tex
\section{Proposed Method}
\label{sec:method}

We use a one-shot learning approach simliar to metric learning based methods~\cite{koch2015siamese,kim2018co,snell2017prototypical,vinyals2016matching}, which learn an embedding space as general as possible by means of a metric comparison.
Such approaches consist of two steps: 1) a training step to learn the embedding space with massive data (generic prior knowledge), and 2) a test step involving NN classification with embeddings of novel class data and their support set.
This approach assumes that the data used in the training step is unrelated to the class of the test phase but has a distribution similar to that of the test data.
Moreover, the embedding is expected to be informative so that one to five support samples (one-shot to few-shot) for each novel class can be sufficiently generalized.

The Variational prototyping-encoder (VPE) differs from metric learning in terms of how it induces a generalized embedding space. 
Instead of determining a user-selected metric to induce an embedding space, VPE learns a generative model with a continuous distribution of data.
VPE seeks the embedding space via a meta-task; conditional image translation from a real image to a prototype.
Additionally, VPE guides distribution learning using prior information about prototypes.

In this paper, we denote a scenario with a support set consisting of $C$ classes with $K$ samples per class as $C$--way $K$--shot classification.
We assume that a single prototype ($K{=}1$) is given for each class as a supported sample, \ie, one-shot classification with a single prototype.

\subsection{Variational Prototyping-Encoder}

Let us consider a paired dataset $\calX = \{ (\bx, \bt)^{(i)}\}_{i=1}^N$, where 
$\bx$ is the real image sample, $\bt$ denotes its corresponding prototype image, and we assume respective i.i.d. samples.
In our scenario, each class has only a single prototype $\bt$ which acts as a label. 
We assume a data generation process similar to a variational autoencoder (VAE)~\cite{kingma2013auto}, but the generated target value is not data $\bx$ but $\bt$: \ie, a latent code $\bz^{(i)}$ is generated from a prior distribution $p_\theta(\bz)$, after which a prototype $\bt^{(i)}$ is generated from a conditional distribution $p_\theta(\bx|\bz)$.
Because this process is hidden, the parameter $\theta$ and the latent variables $\bz^{(i)}$ are unknown.
Thus, we approximate the inference by means of a variational Bayes method.

The parameter approximation is done via marginal likelihood maximization.
Each log marginal likelihood of the individual prototype $\log p_\theta(\bt^{(i)})$ can be lower bounded by 
\begin{equation}
\begin{aligned}
\log p_\theta(\bt) &= \log\int_\bz p_\theta(\bt, \bz) = \log \int_\bz p(\bt, \bz) \tfrac{q_\phi(\bz|\bx)}{q_\phi(\bz|\bx)} \\
& = \log\left( \mathop{\mathbb{E}}\nolimits_{q_\phi(\bz|\bx)} \tfrac{p(\bt, \bz)}{q_\phi(\bz|\bx)} \right)\\
& \geq
\mathop{\mathbb{E}}\nolimits_{q_\phi(\bz|\bx)} \left[
\log {p_\theta(\bt, \bz)}  - \log {q_\phi(\bz|\bx)}
\right]\\
& \textrm{\quad (by Jensen's inequality)}\\
& = \mathop{\mathbb{E}}\nolimits_{q_\phi(\bz|\bx)} \left[
\log {p_\theta(\bt| \bz)} \right]
- D_{KL} \left[ {q_\phi(\bz|\bx)}||{p_\theta(\bz)} \right],
\end{aligned}
\label{eq:elbo}
\end{equation}
where $D_{KL}[\cdot]$ is the Kullback-Leibler (KL) divergence, and a proposal distribution $q_\phi(\bz|\bx)$ is introduced to approximate the intractable true posterior.  
The distributions $q_\phi(\bz|\bx)$ and $p_\theta(\bt| \bz)$ are termed a probabilistic encoder and decoder (or a recognition model and a generative model) respectively.
By maximizing the variational lower bound in \Eref{eq:elbo}, we can determine the model parameters $\phi$ and $\theta$ of the encoder and decoder.

\Eref{eq:elbo} is different from the VAE~\cite{kingma2013auto}.
The VAE is derived from the marginal likelihood over the input data $\bx$, and its lower bound models the self-expression of the input, as 
\begin{equation}
\log p_\theta(\bx) \geq \mathop{\mathbb{E}}\nolimits_{q_\phi(\bz|\bx)} \left[
\log {p_\theta(\bx| \bz)} \right]
- D_{KL} \left[ {q_\phi(\bz|\bx)}||{p_\theta(\bz)} \right].
\end{equation}
In this formulation, $\bx$ is encoded to $\bz$ and reconstructed from $\bz$, while our method encode the input $\bx$ to $\bz$ and translate to a prototype $\bt$ like image-to-image translation~\cite{isola2017image}.
Since prototypes are on a canonical domain with canonical color without perturbation in real objects, our method translates real image inputs to the corresponding prototypical images invariant to real-world perturbations such as background clutter, geometric and photometric perturbations.
In this sense, VPE is related to the denoising autoencoder~\cite{vincent2010stacked,bengio2013generalized} in that VPE acts as a real-world perturbation normalization and may result in embeddings (latent $\bz$) invariant or robust to the perturbations.

In order to efficiently train the parameters by stochastic gradient descent (SGD), we follow Kingma and Welling~\cite{kingma2013auto} to derive a differentiable surrogate objective function by assuming Gaussian latent variables and drawing samples $\{\bz^{(s)}\}_{s=1}^S$ from $q_\phi(\bz|\bx)$.
The empirical loss is then derived as follows:
\begin{equation}
\calL(\bx, \bt;\theta, \phi) {=} 
\frac{1}{S}\sum\limits_{s=1}^S -\log p_\theta(\bt|\bz^{(s)}) + {D_{KL}}[q_\phi(\bz|\bx)\parallel p_\theta(\bz)].
\label{eq:loss}
\end{equation}
The reparameterization trick~\cite{kingma2013auto} is used for \Eref{eq:loss} to be differentiable, whereby $q_\phi(\bz|\bx)$ is re-parameterized with a neural network $g_\phi(\cdot)$, \ie, $\bz^{(s)}$ is sampled by $\bz^{(s)} = g_\phi(\bx^{(i)},\bepsilon^{(s)}) = \bmu^{(i)} {+} \bsigma^{(i)}{\odot}\bepsilon^{(s)}$, where $\bepsilon\sim \calN(\mathbf{0}, \bI)$ and $\odot$ denotes element-wise multiplication.
In addition, the decoder $p_\theta(\bt|\bz)$ is modeled by a  neural network.
We can efficiently minimize \Eref{eq:loss} by SGD with a mini-batch.

In \Eref{eq:loss}, the first and second term correspond to the reconstruction error and distribution regularization term respectively. KL divergence regularizes the latent space by encouraging the distribution of $\bz$ follows the prior distribution, which prevents the distribution from collapsing while 
mapping similar data inputs to nearby locations in the latent space.
Furthermore, the loss induces the mapping of various real images to a single prototype image of the same class. 
This enables the distribution of the latent vectors of real images within the same class to be encapsulated by conditioning its prototype.

For the reconstruction loss in \Eref{eq:loss}, 
any reconstruction loss can be used, from basic losses ($\ell_{1}$- and $\ell_{2}$-norm) to advanced losses (perceptual loss~\cite{hou2017deep} and generative adversarial loss \cite{goodfellow2014generative,larsen2016autoencoding}).
We used the simple binary cross entropy (BCE) loss with real valued targets in $[0,1]$, finding that it is sufficiently efficient for prototypes because many prototypes consist of primary colors within the range of $[0,1]$.
More exploration of loss functions will lead to improvement.

{\flushleft \bf Test phase.}
The learned encoder is only used as a feature extractor.
Given a novel class support set of prototypes, we initially extract their features from the encoder and store them in the support set, (one-shot learning).
Subsequently, when an input query is given, we extract its feature by the encoder and classify by NN classification by retrieving the support set (\Fref{fig:VAE_oneshot}).
Because we assume Gaussian latent variables, we can measure the similarity by Euclidean or Mahalanobis distances.
In this work, we simply use the Euclidean distance for NN classification. We leave the development of advanced metrics as a future work.

{\flushleft \bf Comparison with other approaches.}
In classification, metric learning based one-shot methods~\cite{koch2015siamese,kim2018co,snell2017prototypical,vinyals2016matching} learn non-linear mappings suitable for the given metric distances with labels.
Label information groups data based on discrete decisions as to whether samples belong to the same class or not. 
This tends to be discriminative for the seen classes.
However, it would be difficult to expect the features of images from unseen classes to be distributed meaningfully over the feature space learned in such a manner.\footnote{We compare t-SNE visualizations of several metric learning approaches in the supplementary material offering support of this claim.}  Therefore, several methods have attempted to alleviate the shortage of the metric loss, such as multiple pairwise regularization \cite{kim2018co} and attentional kernel with conditional embedding \cite{vinyals2016matching}, but still limited.

Without directly fixing a metric, our model learns an embedding space in a wholly different manner.
VPE with the prototype reconstruction loss learns the meta-task of normalizing real images and indirectly learns the relative similarities of real images as well as latent features according to the degree of appearance similarity with the corresponding prototypes.
We will show in the experimental section that learning appearance similarity in the image domain allows better generalization.

\subsection{Network architecture}
\label{sec:architecture}

We build an encoder with three convolution layers followed by one fully connected layer each for mean and variance predictions. Each convolution layer has a stride size of 2, downsizing the feature map by a factor of 2. Every convolution layer is followed by batch normalization and leaky ReLU. The final layer is a fully connected layer converting a feature map into a predefined latent variable size. The convolution filter size and latent variable size follow that of the Idsia network~\cite{cirecsan2012multi} which has been the best traffic sign classification network within the GTSRB benchmark~\cite{stallkamp2012man}. Layers of the decoder are in an inverse order of the encoder layers; \ie, a fully connected layer followed by three convolution layers. We upsample by a factor of 2 before each convolution to recover the feature size to the original input size. All convolution kernels in the decoder are set to 3 $\times$ 3. As in the encoder, every convolution in the decoder is followed by batch normalization and leaky ReLU.

\subsection{Data augmentation}

We apply random rotation and horizontal flipping to both the real images and prototypes identically to train our networks. 
Augmentation diversifies the training samples including the prototypes.
We can easily imagine that a sign with the right directional arrow can become an arrow sign with any directional form after augmentation. This helps the generalization of our network, and we observed that it improves the performance noticeably, whereas it does so subtly in other metric learning methods.

%% file: 04_Experiment.tex
\section{Experiment}

In this section, we first describe the data set configuration and the overall experiment setup, and then implementation details.
We compare the following methods for one-shot classification and retrieval tasks: Siamese networks~\cite{koch2015siamese} (SiamNet), 
Quadruplet networks~\cite{kim2018co} (QuadNet), 
Matching networks~\cite{vinyals2016matching} (MatchNet)
and the proposed networks (VPE).
We also present additional qualitative analyses, t-SNE visualization, a distance heat map between prototypes and real images, and prototype reconstruction.

\begin{table}[h]
		\centering
		    \vspace{-0mm}
		\resizebox{1\linewidth}{!}{
			\begin{tabular}{lccccc}
				\toprule
                Dataset & GTSRB  & TT100k & BelgaLogos & FlickrLogos-32 & TopLogo-10\\ 
				\midrule
				Instances& 51,839 & 11,988  & 9,585 & 3,404  & 848 \\ 
                Classes  & 43    & 36     & 37   & 32    & 11 \\
				\bottomrule
			\end{tabular}
		}
	\vspace{0mm}
	\caption{Symbol dataset specifications.}
	\label{tbl:dataset}
    \vspace{-4mm}
\end{table}
{\flushleft \bf Datasets and experiment setup.}
The evaluation is conducted on two traffic sign datasets and three logo datasets with different training and test set selections. The size and number of classes for each dataset are described in \Tref{tbl:dataset}.
For detailed explanations about the datasets and more image visualizations, please refer to the supplementary material. 

To validate our one-shot learning method, we perform a cross-dataset evaluation by separating the training and test datasets, which is a more challenging setup compared to the use of splits within a single dataset. 
We denote `All' for evaluating the entire dataset and `Unseen' for evaluating the dataset excluding the classes contained in a training set.
The dataset on the left side of an arrow is used as a training set while that on the right side of an arrow is used as a test set (\Tref{tbl:nn_ts} and \Tref{tbl:nn_logo}), \eg, GTSRB$\rightarrow$TT100k.

\begin{table}[!t]
		\centering
	
		\resizebox{0.925\linewidth}{!}{
			\begin{tabular}{lccc}
				\toprule
                 & \scriptsize GTSRB & \multicolumn{2}{c}{\scriptsize GTSRB}\\ 
                 & \scriptsize $\to$ GTSRB & \multicolumn{2}{c}{\scriptsize $\to$ TT100k}\\ 
                \cmidrule(l{2pt}r{2pt}){2-2} \cmidrule(l{2pt}r{2pt}){3-4}
        Split     & Unseen & All  & Unseen   \\ 
     No. classes& 21 & 36 &  32  \\ 
No. support set    & (22+21)-way & \multicolumn{2}{c}{36-way} \\
				\midrule
				SiamNet \cite{koch2015siamese}		& 22.45 & 22.73 & 15.28  \\ 
                SiamNet{+}\textsf{aug} 				& 33.62 & 28.36   & 22.74  \\ 
				QuadNet* \cite{kim2018co}			& 45.2* & 42.3* & N/A    \\
				MatchNet \cite{vinyals2016matching}	& 26.03 & 53.16 & 49.53  \\ 
				MatchNet{+}\textsf{aug}				& 53.30 & 62.14 & 58.75  \\
				\midrule
                VPE \textsf{(48x48)} 	  	      & 55.30 & 52.08 & 49.21  \\
                VPE{+}\textsf{aug}  		      & 69.46 & 66.62 & 63.91  \\
                VPE{+}\textsf{aug}+\textsf{stn}   & 74.69 & 66.88 & 64.07  \\
                \midrule
				VPE \textsf{(64x64)}              & 56.98 & 55.58 & 53.04  \\ 
				VPE{+}\textsf{aug}                & \bbold{81.27} & \bbold{68.04} & \bbold{64.80}  \\ 
                VPE{+}\textsf{aug}+\textsf{stn}   & \rbold{83.79} & \rbold{73.98} & \rbold{71.80}  \\
                \midrule
                \midrule
                VAE								  & 20.67 & 33.14 & 29.04 \\
                VAE{+}\textsf{aug}				  & 22.24 & 32.10 & 27.98 \\
				\bottomrule
			\end{tabular}
		}
	\vspace{2mm}
	\caption{One-shot classification (Top 1-NN) accuracy ($\%$) on traffic sign datasets. The numbers marked with ``*'' are quoted from their papers. VPE on two different input resolutions, $48\times48$ and $64\times64$, are reported for the evaluations. The best accuracy is marked in blue, and the second best is shown in sky blue.}
        	\label{tbl:nn_ts}
    \vspace{-0mm}
\end{table}

\begin{table}[!t]
		\centering
        \resizebox{1.0\linewidth}{!}{
			\begin{tabular}{lcccc}
				\toprule
				 & \multicolumn{2}{c}{\scriptsize Belga} & \multicolumn{2}{c}{\scriptsize Belga} \\
                 & \multicolumn{2}{c}{\scriptsize $\to$ Flickr32} & \multicolumn{2}{c}{\scriptsize $\to$ Toplogos}\\
                \cmidrule(l{2pt}r{2pt}){2-3} \cmidrule(l{2pt}r{2pt}){4-5} 
		Split  & All  & Unseen & All & Unseen \\
	No. classes	& 32 & 28 & 11 & 6 \\     
No. support set & \multicolumn{2}{c}{32-way} & \multicolumn{2}{c}{11-way} \\

				\midrule
				SiamNet \cite{koch2015siamese}		& 23.25 & 21.37 & 37.37 & 34.92  \\ 
                SiamNet + \textsf{aug} 				& 24.70 & 22.82 & 30.84 & 30.46 \\
				QuadNet \cite{kim2018co}			& 40.01 & 37.72 & 39.44 & 36.62  \\
                QuadNet + \textsf{aug}          	& 31.68 & 28.55 & 38.89 & 34.16  \\
				MatchNet \cite{vinyals2016matching}	& 45.53 & 40.95 & 44.35 & 35.24 \\ 
                MatchNet+\textsf{aug}           	& 38.54 & 35.28 & 28.46 & 27.46 \\ 
				\midrule
				VPE	\textsf{}                   	& 28.71 & 27.34 & 28.01 & 26.36 \\ 
				VPE+\textsf{aug}                	& \bbold{51.83} & \bbold{50.25} & \bbold{47.48} & \bbold{41.82} \\ 
                VPE+\textsf{aug}+\textsf{stn}   	& \rbold{56.60} & \rbold{53.53} & \rbold{58.65} & \rbold{57.75} \\
                \midrule
                \midrule
                VAE								& 25.01 & 25.48 & 21.90 & 15.89 \\
                VAE{+}\textsf{aug}				& 27.17 & 27.31 & 23.30 & 18.59 \\
				\bottomrule
			\end{tabular}
		}
			\vspace{0mm}
	\caption{One-shot classification (Top 1-NN) accuracy ($\%$) on logo datasets. The best accuracy is marked in blue and the second best is shown in sky blue.}
	\label{tbl:nn_logo}
    \vspace{-4mm}
\end{table}

\begin{figure*}[!t]
	\centering	{\includegraphics[width=0.95\linewidth]{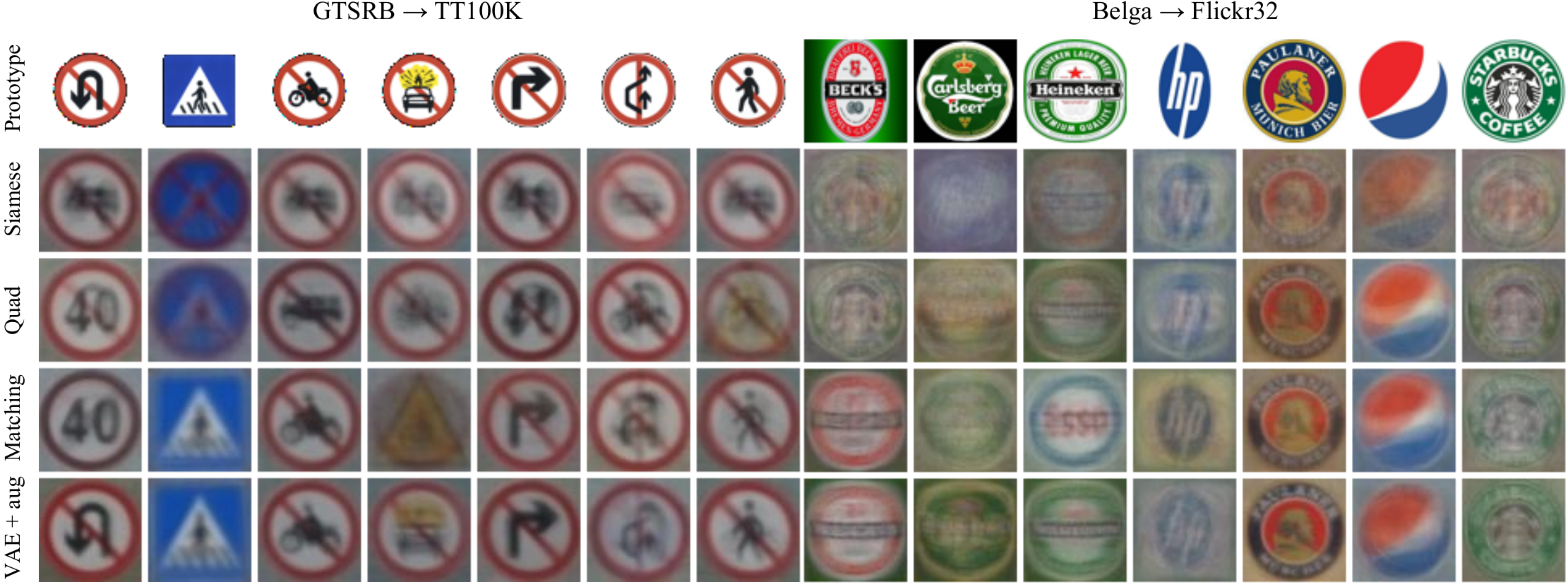}}\hspace{2mm}
	\caption{Average image of top 100 images retrieved by querying prototypes. A clearer image represents a higher retrieval performance.
    The classes shown are selected from unseen classes.}
	\label{fig:mean_image}
	\vspace{-2mm}
\end{figure*}

For logo classification, BelgaLogos~\cite{belgalogos09,letessier2012scalable}, FlirckrLogos-32~\cite{romberg2011scalable} and TopLogo-10 \cite{su2017deep} are used. 
BelgaLogos is used as a training set and remaining datasets are used as the test and validation sets.
For example, in the Belga$\to$Flickr32 case, TopLogo-10 is used as a validation set. 
BelgaLogos and FlickrLogos-32 share four common classes, and BelgaLogos and Toplogo-10 share five common classes. 
We exclude the common classes in the ``Unseen'' test.
For traffic sign classification, the GTSRB~\cite{stallkamp2012man} and TT100K~\cite{zhu2016traffic} datasets are used. 
For the GTSRB$\to$TT100k scenario, we train the model on GTSRB and report the best accuracy tested on TT100K. GTSRB and TT100K shares four common classes.

While the entire dataset is used for training and testing during the cross-dataset evaluation, the GTSRB experiment is performed using only the GTSRB dataset with splits. 
Among a total of 43 classes in GTSRB, we select 22 classes as seen and the remaining 21 classes as unseen. 
GTSRB has two data partitions: the train and test partitions. We trained a model with the training set of the 22 seen classes and evaluate the performance on the test set of all 43 classes. 
The 21 unseen class samples in the training set are used for validation. 
This scenario is unique in that the support set contains all of the seen and unseen prototypes. Because the random chance accuracy of this case becomes far lower, this is a more difficult setup than the typical one-shot evaluation scenario, where a support set is assumed to contain only unseen samples. 
In this setup, we can determine whether a model is biased toward seen classes. 
The details of the GTSRB experiment setup follow the work of Kim \etal~\cite{kim2018co}.

\begin{figure*}[ht!]
\centering
\setlength{\tabcolsep}{1pt}
\footnotesize


\resizebox{0.99\linewidth}{!}{%
\begin{tabular}{cccc}
  \includegraphics[width=0.25\linewidth]{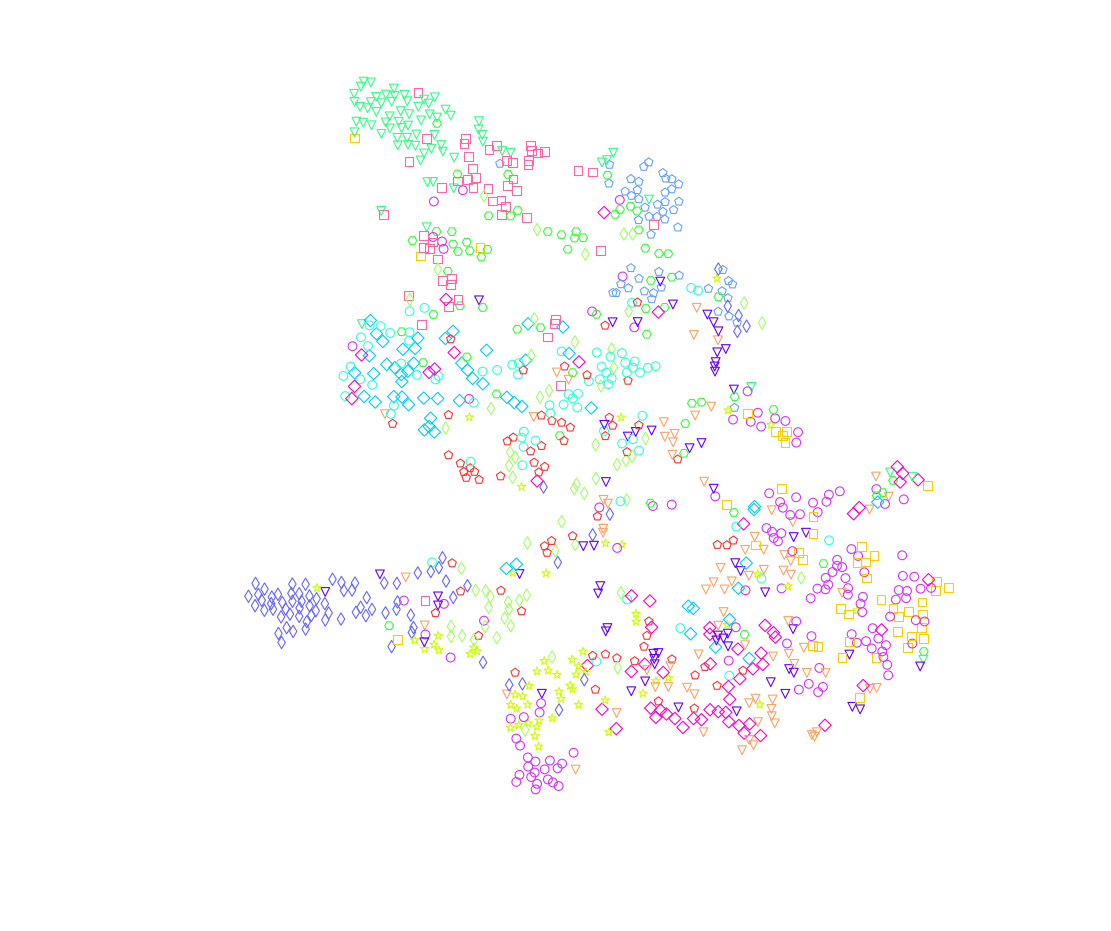} &
  \includegraphics[width=0.25\linewidth]{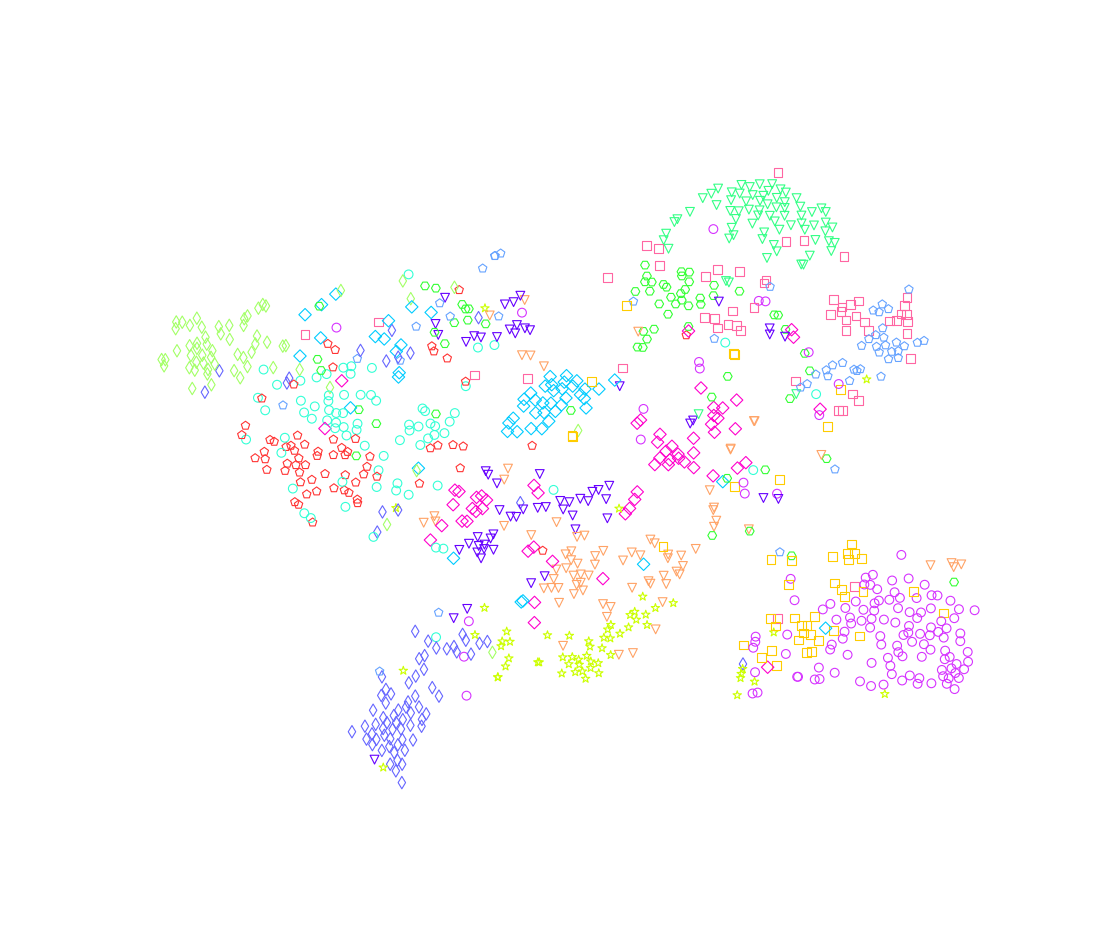} &
  \includegraphics[width=0.25\linewidth]{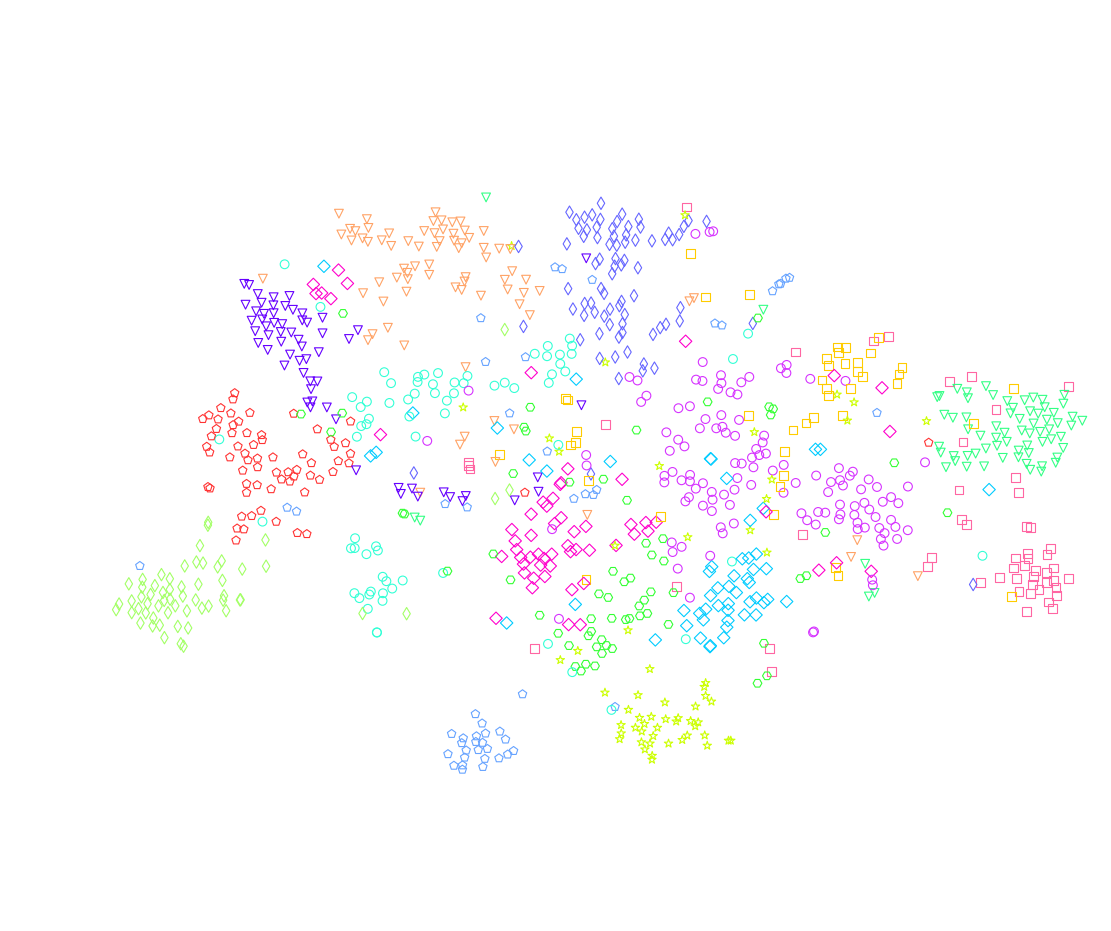} &
  \includegraphics[width=0.25\linewidth]{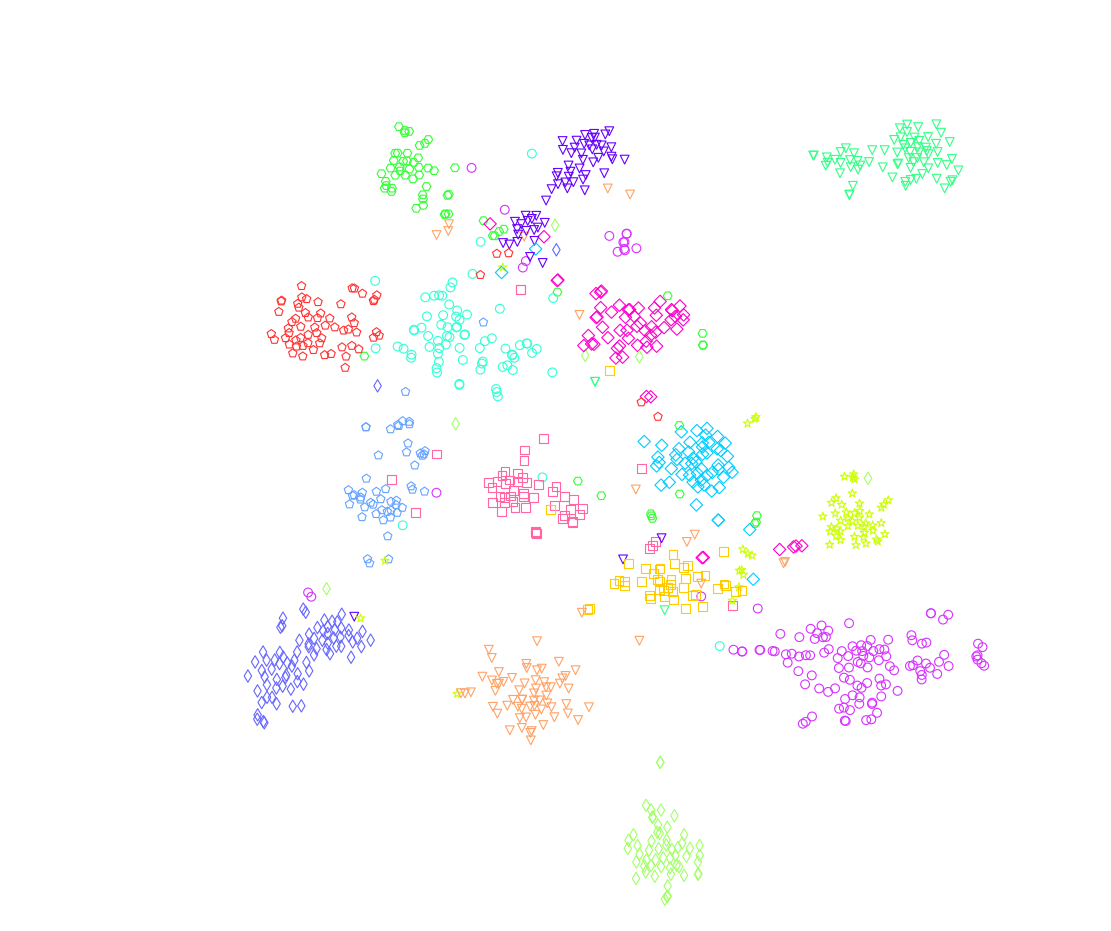}\\
  SiamNet & QuadNet & MatchNet & VPE + aug \\
\end{tabular}
}
\vspace{2mm}
   \caption{t-SNE visualization of features. Features are randomly sampled from 15 unseen classes of Belga$\to$Flickr32 scenario.}
\label{fig:tsne}
\vspace{-4mm}
\end{figure*}

\vspace{-2mm}
{\flushleft \bf Implementation details.}
For a fair comparison, all of the methods in this experiment use IdsiaNet~\cite{cirecsan2012multi} as a base network. We tune to obtain the best performance of the methods, and we use the ADAM optimizer~\cite{kingma2014adam} with a learning rate of $10^{-4}$, $\beta = (0.9, 0.999)$, $\epsilon = 10^{-8}$ and a mini-batch size of $128$ to train the networks.  
The original implementations of SiamNet and MatchNet\footnote{MatchNet implementation are based on, \url{https://github.com/gitabcworld/MatchingNetworks}} are designed for character classification; hence, a base network change is necessary. 
We found that the substitution of the base networks significantly improved the performance outcomes. 
We use input sizes of 48$\times$48 for traffic sign data and 64$\times$64 for logo data but also test different resolution effects as a short ablation study, as shown in \Tref{tbl:nn_ts}. 
The input dimension of the first fully connected layer is adjusted according to the input size so that the final dimension of embedding is fixed at 300 for all methods regardless of the input size. 
The rationale behind a larger size for logos is their various aspect ratios. 
We maintain the aspect ratio by resizing a larger axis of an image to fit the network input size with zero padding.

We also found that SiamNet performs very poorly when trained using prototypes as a query. Therefore, we trained SiamNet using only real images for both query and positive, negative sample pairs.
QuadNet is reproduced using IdsiaNet and is evaluated on the logo datasets. 
However, the original implementation fusing two Siamese networks performed poorly on logos. 
We modified QuadNet to share all of the parameters of the networks in order to stabilize the training instead of using two Siamese networks. 
We conjecture that the failure of the original implementation on logos stems from a quality of the training set. 
GTSRB is larger than logo datasets containing samples of a higher quality, whereas logo datasets have fewer samples, and some images are severely distorted, including non-rigid transformations, \eg, logos printed on curved bottles or wrinkled clothes. 

The term \emph{aug} represents the random flip and rotation augmentation applied, and \emph{stn} is a spatial transformer~\cite{jaderberg2015spatial} attached to the encoder part, \ie, the improved IdsiaNet suggested by the Moodstock team.\footnote{Their experiment achieved a meaningful performance improvement of IdsiaNet on the traffic sign classification. For more detail, please refer to, \url{https://github.com/moodstocks/gtsrb.torch}}
For the \emph{stn} version, the spatial transformer modules are applied before the $1st$ and $3rd$ convolution layers in the encoder part.
By doing this, we can show that the proposed method has the potential to be improved further if advanced techniques are adopted.
Prototype images and real images are randomly sampled at a 1:200 ratio during training.



\subsection{One-shot classification (Real to prototypes)}
\label{sec:one-shot}

The one-shot classification performances are reported in \Tref{tbl:nn_ts} and \Tref{tbl:nn_logo}. 
VPE and its variants perform better than competing approaches in most cases. 
The margin is significant in the traffic sign task while less of an improvement was noted on logo datasets. 
We surmise that this performance gap comes from the quality of the training dataset. 
As mentioned earlier, GTSRB is the largest dataset among the five datasets, and traffic sign images are well localized with consistent aspect ratios, whereas logos are more challenging due to various aspect ratios, color variations, and non-rigid deformation.

Interestingly, the augmentation improves VPE noticeably, though it has less of an effect with the other approaches. 
A possible explanation for this tendency is that VPE learns a pseudo image transform process and tends to measure a type of perceptual similarity which is less sensitive to subtle input changes.
This would not be the case with direct metric learning methods, as subtle perceptual changes such as flipping in the input domain do not have to be mapped to similar embedding vectors. Refer to the distance heat map shown in \Fref{fig:dist_mat}


We emphasize the GTSRB scenario, of which the support set used in the test phase involves seen classes during training as well as unseen novel classes.
This allows us to measure overfitting to seen classes. 
This is an evaluation different from typical one-shot classification setups, where a support set does not contain any samples from training classes, making the process far easier.
In this scenario, MatchNet shows poor performance without augmentation. We conjecture that this is due to the attentional kernel, which is biased to favor seen classes. 

The VAEs in Tables~\ref{tbl:nn_ts} and \ref{tbl:nn_logo} are models that share the same architecture with our VPE, but trained with variational auto-encoding loss~\cite{kingma2013auto} without prototypes. 
It is reported as a reference to show how VAE performs without prototype learning. 
The low performance of VAE has two possible causes: 
1) the lack of supervision to reduce the domain gap between the real and prototype domains, and
2) the lack of explicit information to induce clustering effects according to actual classes, which makes the VAEs difficult to adjust which level they should cluster or distinguish across samples.



\subsection{Image retrieval test (Prototypes to real)}

\begin{table}[!t]
	\centering
	\resizebox{0.99\linewidth}{!}{
    	\begin{tabular}{l@{\hskip 3mm}c@{\hskip 3mm}c@{\hskip 3mm}c@{\hskip 3mm}c}
    		\toprule
             & \scriptsize GTSRB & \scriptsize GTSRB & \scriptsize Belga & \scriptsize Belga\\ 
            AUC & \scriptsize $\to$ GTSRB & \scriptsize $\to$ TT100k & \scriptsize $\to$ Flickr32 & \scriptsize $\to$ Toplogos \\ 
            \midrule
    		SiamNet                             & 8.75  & 4.83  & 20.56 & 18.13  \\ 
            Quadnet 	                        & \textsf{n/a}   & \textsf{n/a}   & 32.40 & 20.51  \\ 
    		MatchNet          		            & 57.99 & 41.00 & 44.47 & 46.13  \\
    		VPE{+}\textsf{aug}                  & \bbold{64.77} & \bbold{41.79} & \bbold{48.61} & \bbold{49.39}  \\ 
    		VPE{+}\textsf{aug}+\textsf{stn}     & \rbold{85.29} & \rbold{64.04} & \rbold{63.87} & \rbold{70.22}  \\
    		\bottomrule
    	\end{tabular}
	}
	\vspace{1mm}
	\caption{AUC score of retrieval experiments. }
    \label{tbl:retrieval}
    \vspace{-5mm}
\end{table}

Average image~\cite{oliva2007role,zhu2014averageexplorer} can provide an intuitive visual understanding of multiple images.
In this experiment, we summarize image retrieval results using average images. With the trained one-shot models, by querying prototypes, images are retrieved based on the metrics of each method. An average of the retrieved image qualitatively visualizes the discriminative power of the learned embeddings of the models. 
A fine average image is obtained only if there are negligible outliers in the retrieved results. We provide average images by retrieval along with prototypes for comparison (\Fref{fig:mean_image}). The result clearly shows that VPE is effective for a comparison in the opposite direction, \ie, prototype $\to$ real images.

While average images provide qualitative measure of the retrieval task, we also report the quantitative retrieval performance in \Tref{tbl:retrieval} using the area under the precision-recall curve (AUC). The relative retrieval performance between the competing approaches are similar to that of the one-shot experiments (Sec. \ref{sec:one-shot}).



\subsection{Additional analyses}

\begin{figure}[b]
\centering
{\includegraphics[width=1.0\linewidth]{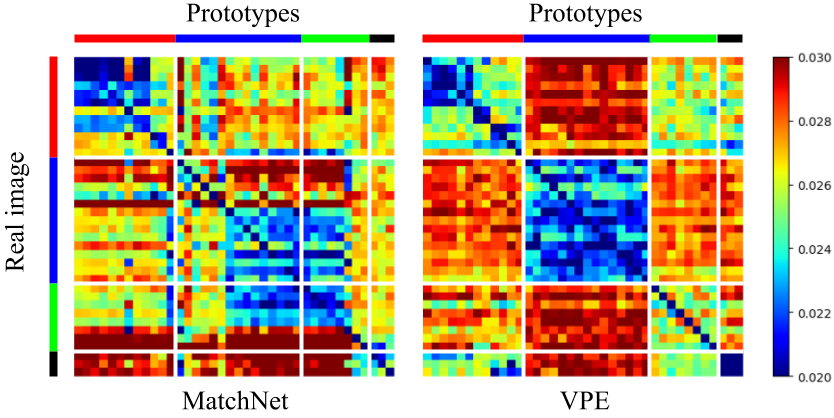}}\hspace{2mm}
	\vspace{-3mm}
	\caption{ Average distances between real images and prototypes from GTSRB scenario are visualized as heatmap matrices. }
	\label{fig:dist_mat}
	\vspace{-0mm}
\end{figure}

\begin{figure*}[!t]
\centering
{\includegraphics[width=0.95\linewidth]{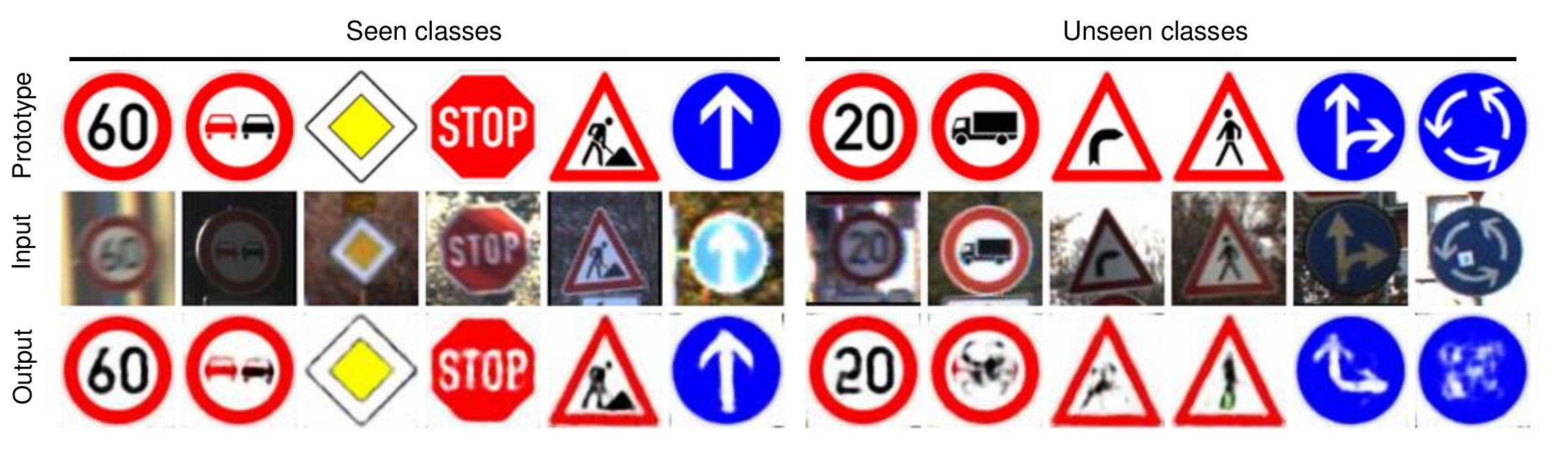}}\hspace{2mm}
	\vspace{-0mm}
	\caption{ The VPE output on GTSRB scenario. }
	\label{fig:recon}
	\vspace{-3mm}
\end{figure*}



{\flushleft \bf Similarity measure.}
One-shot classification focuses on general classification capability including that for unseen classes. Understanding image similarity and dissimilarity is an important capability for one-shot classification. 
Metric-based approaches adopt metric losses induced from labels, semantically coarser information without image level similarity, while the proposed method uses appearance similarity and thus semantically finer information.

To demonstrate the quality of learned image similarity further, we show, in \Fref{fig:dist_mat}, the average distance matrix between real images and prototypes from the GTSRB dataset.
Each column of distance matrices is $l_1$ normalized for visualization purposes.
The GTSRB dataset has 38 classes that are categorized into four groups: Prohibitory, Danger, Mandatory and Others. 
Classes within the same category have a similar external shape while differing in terms of the interior contents. 
Subsequently, we mark the classes of each group with one color along the x-axis and y-axis of the matrices and use red, blue, green and black for the four groups listed above, respectively. 
The diagonal of the matrix represents the distance between corresponding pairs of real images and prototypes.
We compare the distance matrices between MatchNet and the proposed VPE. 
The VPE distance matrix clearly shows a block patterned distance map, indicating that VPE captures appearance similarity in the latent space. 
On the other hand, although MatchNet show short distances along diagonal, there is no clear block pattern aligned with category sets.

{\flushleft \bf Embedding visualization.}
In \Fref{fig:tsne}, We compare t-SNE~\cite{maaten2008visualizing} plots of the embedding spaces of the methods to understand the learned embeddings of unseen data. 
We assign colors according to class labels to observe the discriminative behavior.
VPE shows a clear separation of sample points, whereas the competing approaches show partially mixed distributions. 
This distribution difference is consistent with the results from the one-shot classification experiment. 
It would suggest that the appearance based loss leads to better learning of the general characteristics of symbols as apposed to direct metric losses. 

{\flushleft \bf Prototype reconstruction.}
While the reconstruction task is an auxiliary task for training the proposed VPE networks, for a better understanding of the image translation behavior to unseen data, we visualize the generated outputs in \Fref{fig:recon}.
The model robustly generates prototypes of seen classes regardless of motion blur, illumination variations, or low resolutions.
While the generation performance is not accurate for unseen classes, it still captures some level of the characteristics of these classes in the input images.
It is interesting to note that VPE feasibly handles high-level categories, such as prohibitory (red circle) and danger (red triangle) categories.
Although the fine-details of the symbol contents are not accurate, the locations of the blobs are roughly aligned with the contents in the prototypes. 
This suggests that even the rough generation is still effective for NN classification in the latent space and may apply to a high-level conceptual understanding of novel contexts.

%% file: 05_Conclusion.tex
\section{Conclusion}

We present a new one-shot learning approach based on a generative loss.
The key idea of the proposed VPE invloves the use of reconstruction loss to learn to induce indirect perceptual similarities of real images and their corresponding prototypes, as opposed to the use of a pre-determined metric. 
A prototype reconstruction experiment (\Fref{fig:recon}) demonstrated that our VPE implicitly learns favorable knowledge about how a real image can be neutralized against real-world perturbations, such as radiometric and geometric perturbations.
VPE appears to capture high level prototype concepts from images of unseen classes distorted by real world perturbations to some extent.
This is fundamentally different from metric learning approaches, as they use label information to group available data in the training phase, making it difficult to expect the generalization of similarities to unseen classes. 

We quantitatively and qualitatively validated the performance of the proposed methods on multiple datasets and demonstrated its favorable performance over competing approaches.
Despite the noticeable performance improvement of VPE, it is simple to train and the resulting architecture is simple as well.
In this regard, the principal behind VPE would lead to various applications in the future.



%% file: 06_Supple.tex
\section{Supplementary materials}

Here, we present additional details pertaining to the datasets and experiments that could not be included in the main text due to space constraints. All figures and references in this supplementary
file are self-contained.

The contents included in these supplementary materials are as follows: 1) The network architecture, 2) Detail descriptions of the datasets used, 3) Embedding space visualization, and 4) Qualitative results of image retrieval.

\subsection{Architecture}

\begin{figure}[!h]
	\centering	
	{\includegraphics[width=1\linewidth]{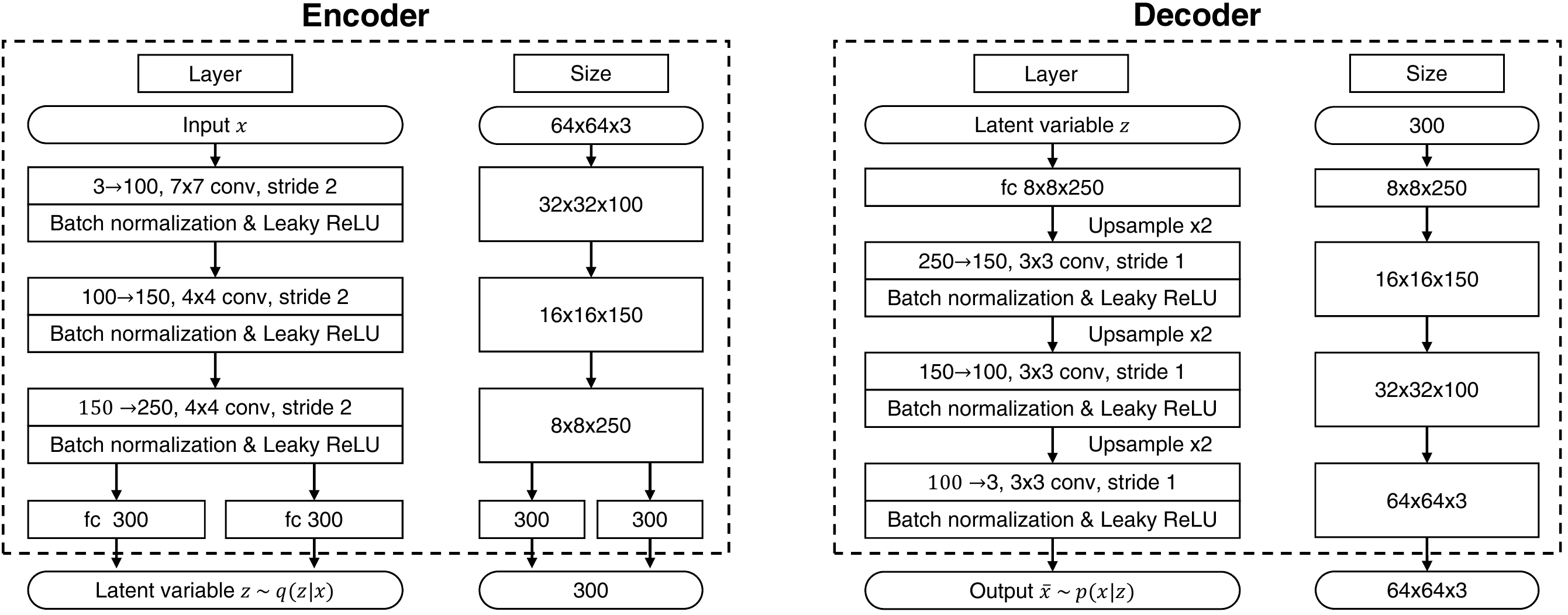}}\hspace{2mm}
	\vspace{-2mm}
	\caption{Architectures specifications of encoder and decoder blocks of the proposed variational autoencoder.}
	\label{fig:architecture}
	\vspace{-1mm}
\end{figure}

The detailed VPE network architecture is shown in \Fref{fig:architecture}.

\clearpage

\subsection{Datasets}

\begin{table}[!h]
	\begin{center}
		\resizebox{0.7\linewidth}{!}{
			\begin{tabular}{cccccc}
				\toprule
                Dataset & GTSRB  & TT100k & BelgaLogos & FlickrLogos-32 & TopLogo-10\\ 
				\midrule
				Instances& 51,839 & 11,988  & 9,585 & 3,404  & 848 \\ 
                Classes  & 43    & 36     & 37   & 32    & 11 \\
				\bottomrule
			\end{tabular}
		}
	\end{center}
	\caption{Symbol dataset specifications}
	\label{tbl:dataset2}
\end{table}

In this section, we present the details of each dataset used for the experiments in the main text. \Tref{tbl:dataset2} is provided to summarize the statistics of the datasets.


\vspace{1mm}\noindent\textbf{GTSRB}
GTSRB \cite{stallkamp2012man} is the largest dataset for traffic-sign recognition. It contains 43 classes categorized into three larger categories: prohibitory, danger and mandatory. The dataset contains illumination variations, blur, partial shadings and low-resolution images as well as imbalanced sample distribution. The training set contains 39,209 images and the test set contains 12,630 images.



\vspace{1mm}\noindent\textbf{TT100K}
Tsinghua-Tencent 100K (TT100K) \cite{zhu2016traffic} is a Chinese traffic sign detection dataset that includes more than 200 classes. We cropped traffic sign
instances from scenes to build a classification dataset. We filtered out instances with side lengths of less than 20 pixels because they are either not recognizable or miss annotated.
Among more the defined classes, the 36 classes are selected for the evaluation that have available corresponding prototypes and a sufficient number of samples. For more details about the TT100K dataset, please refer to the work of Kim \etal \cite{kim2018co}.



\vspace{1mm}\noindent\textbf{FlickrLogos-32 Dataset} 
FlickrLogos-32 \cite{romberg2011scalable} is a collection of images from Flickr containing 32 different logos. Most of the images contain a few and relatively clean, recognizable logo instances located near the center of an image compared to other datasets \cite{belgalogos09,tuzko2017open}.
The dataset is published to evaluate logo detection and recognition systems with 32 logo classes defined. The dataset has a total of 2,240 logo images, and it is partitioned into 10 training images, 30 validation images and 30 test images per class. It also contains 6,000 no-logo images to evaluate the false alarm rates of recognition systems. We cropped logo instances using bounding box annotations to evaluate our classification systems. In total, 3,372 logo instances were gathered by cropping.



\vspace{1mm}\noindent\textbf{BelgaLogos Dataset} 
BelgaLogos \cite{belgalogos09,letessier2012scalable} is composed of 10,000 images from various aspects of everyday life with 37 logo classes annotated in a bounding box format. Unlike FlickrLogos-32, logos appear at diverse locations with large-scale variations, blur, saturation and occlusions. The quality levels of the samples are rated as either `OK' or `Junk' depending how clearly a sample is recognizable by human annotators. We cropped both `OK' and `Junk' logo instances to build a logo classification dataset. In total, 9,475 instances were collected. While FlickrLogos-32 shows an equal sample distribution per class, BelgaLogos shows a severe class imbalance from a small-sized class (2 samples) to a large-sized class (2,242 samples).



\vspace{1mm}\noindent\textbf{TopLogo-10 Dataset}
TopLogo-10 \cite{su2017deep} contains 10 logo classes related to popular cloth, shoes and accessory brands. The images are collected from product images that are relatively clean and recognizable. Each class contains 70 images. We cropped logo instances using bounding box annotations and gathered a total of 853 logo samples. For the experiment, we defined a total of 11 logo classes by separating the `Adidas' class into the `Adidas-logo' and the `Adidas-text' classes.


\subsection{Embedding space}

We provide t-SNE~\cite{maaten2008visualizing} plots using each method introduced in the main text. We select two representative evaluation scenarios, GTSRB$\to$TT100K and Belga$\to$Flickr32, for visualization. The result shows a clear difference between the feature distribution of VPE and the remaining feature spaces. It should be noted that VPE generates a more discriminative feature distribution compared to those by the competing approaches.

\begin{figure*}[h]
\vspace{-1mm}
\begin{center}
\setlength{\tabcolsep}{1pt}
\footnotesize
\resizebox{0.69\linewidth}{!}{%
\large
\begin{tabular}{cc}
  
  \includegraphics[width=0.5\linewidth]{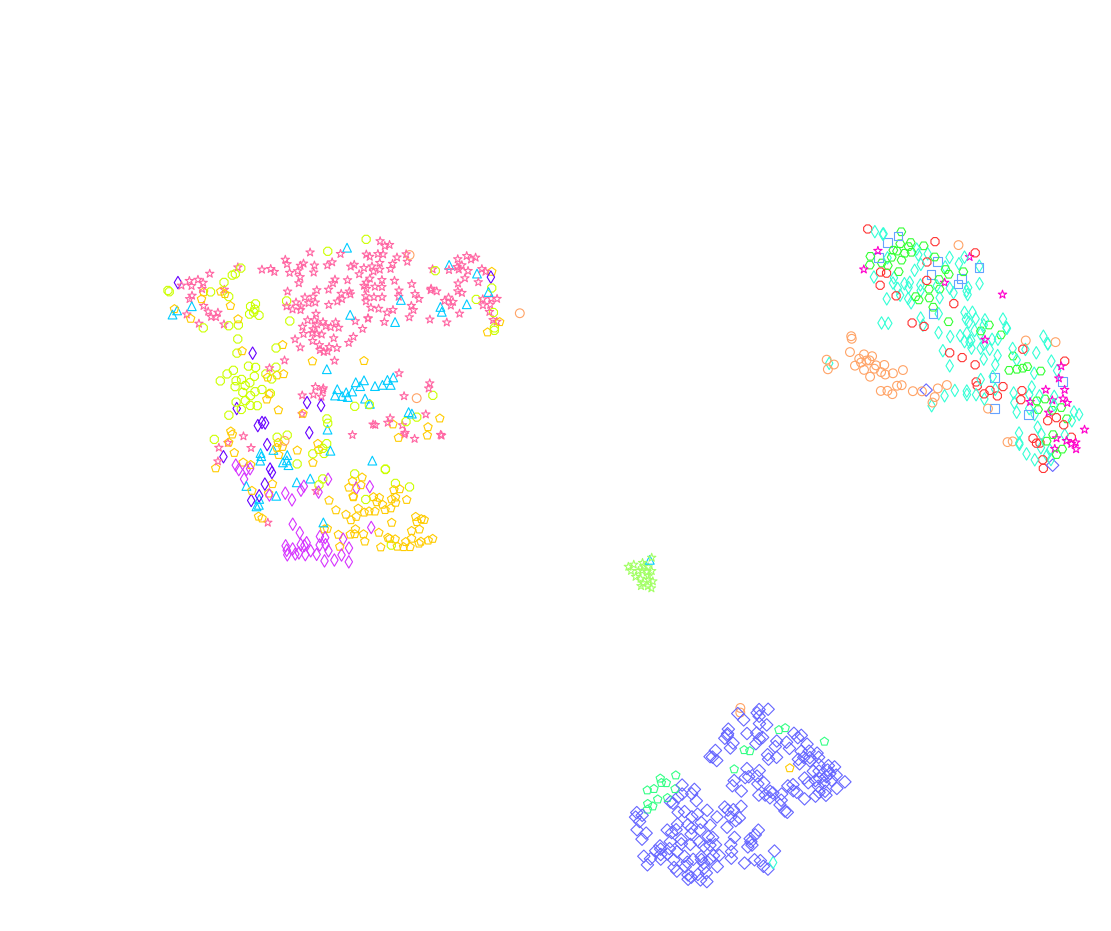} &
  \includegraphics[width=0.5\linewidth]{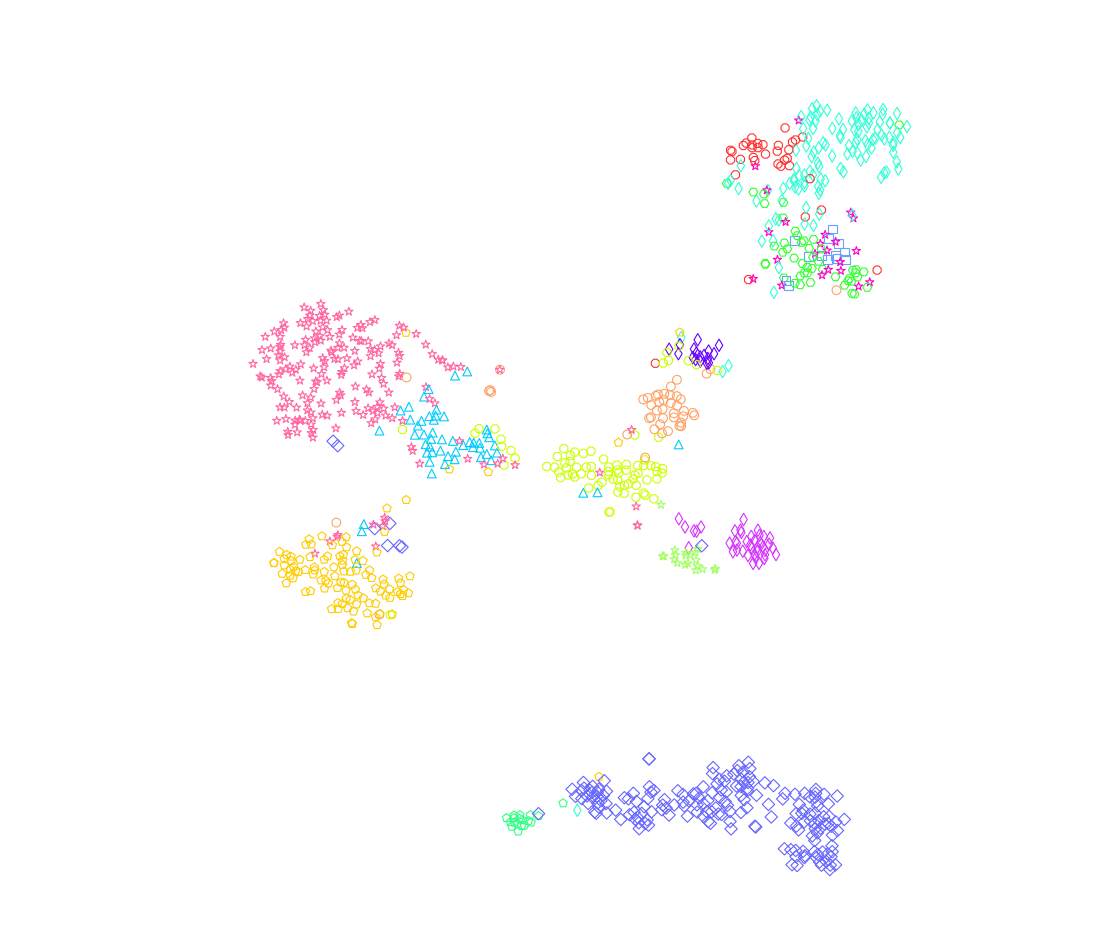} \\
  Siamese networks & Quadruplet networks \\
  
  \includegraphics[width=0.5\linewidth]{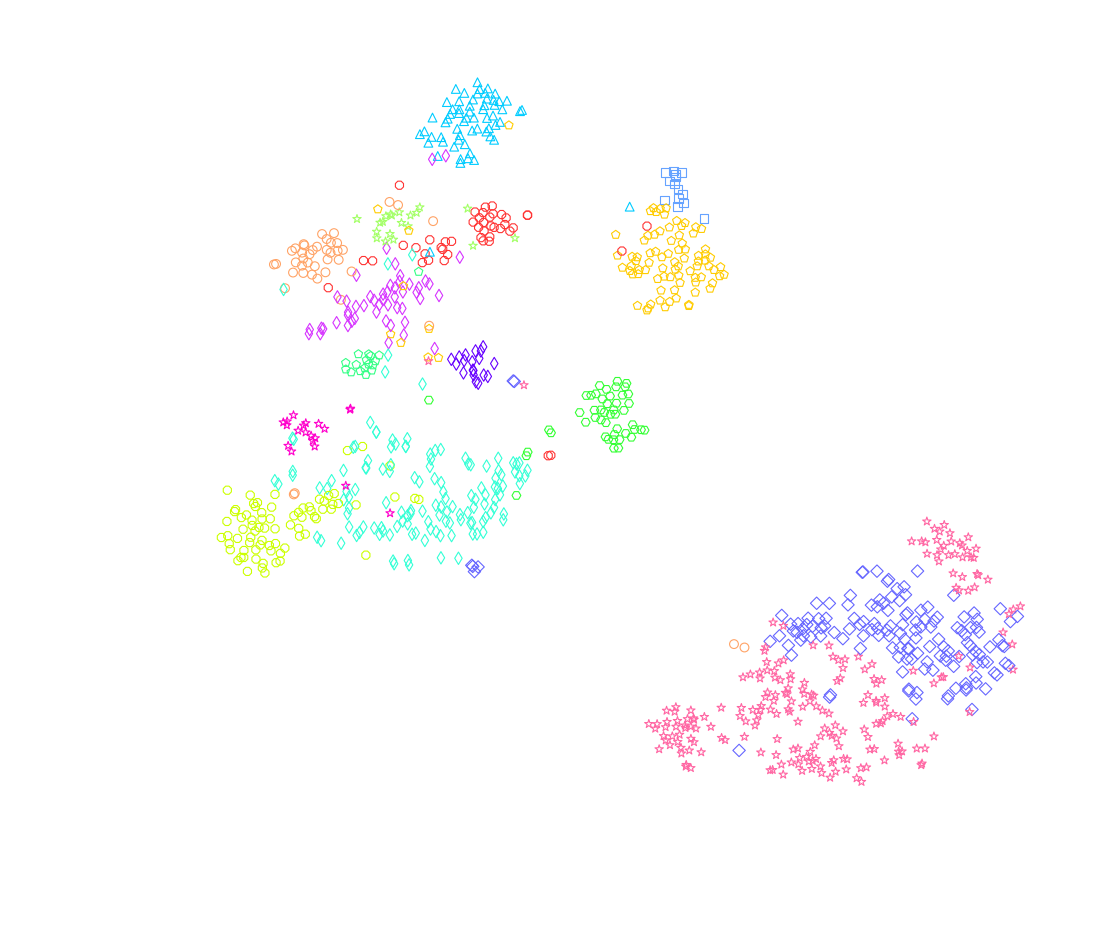} &
  \includegraphics[width=0.5\linewidth]{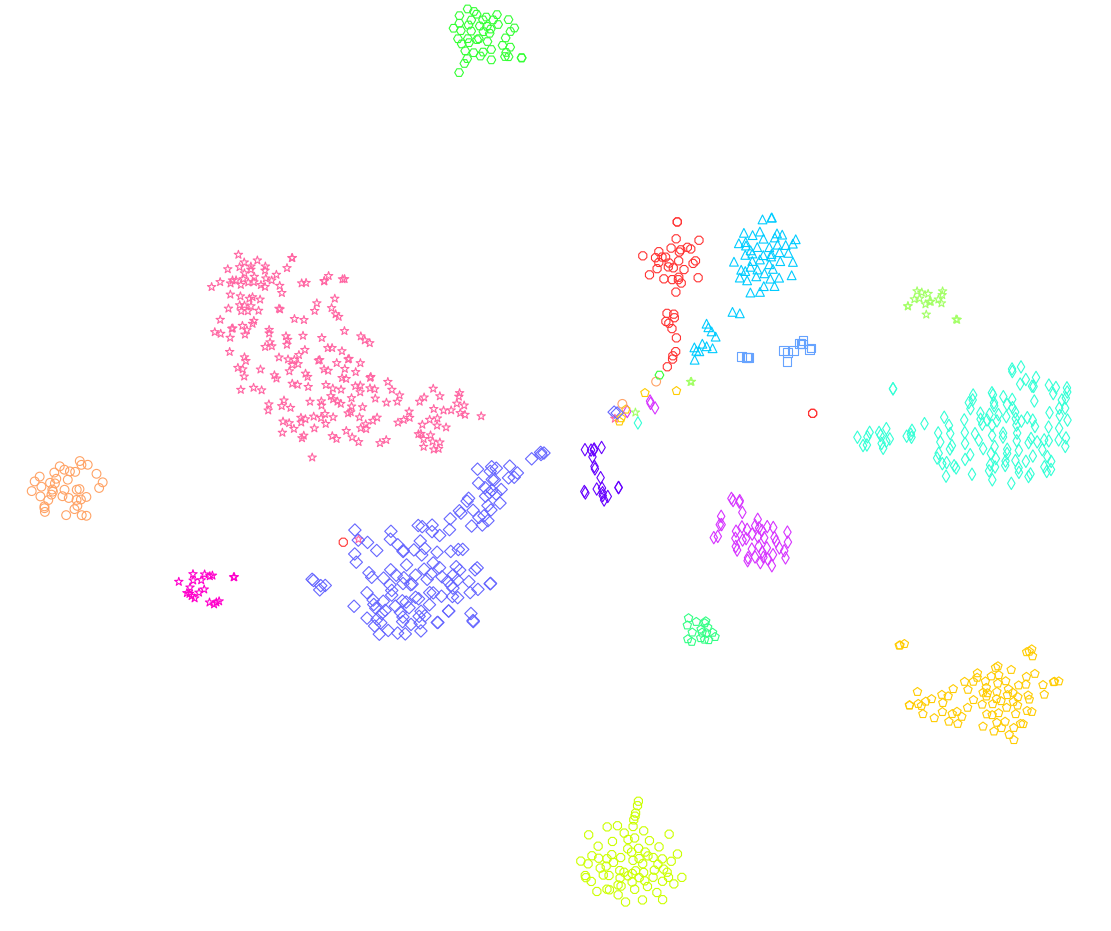}\\
  Matching networks & VPE + aug \\
  
\end{tabular}
}
\end{center}
\vspace{-2mm}
   \caption{t-SNE visualization of features on embedding space. Features are randomly sampled from 15 different unseen classes under the GTSRB$\to$TT100K scenario for visualization.}
\vspace{-2.5mm}
\label{fig:tsne_tf}
\end{figure*}

\begin{figure*}[h]
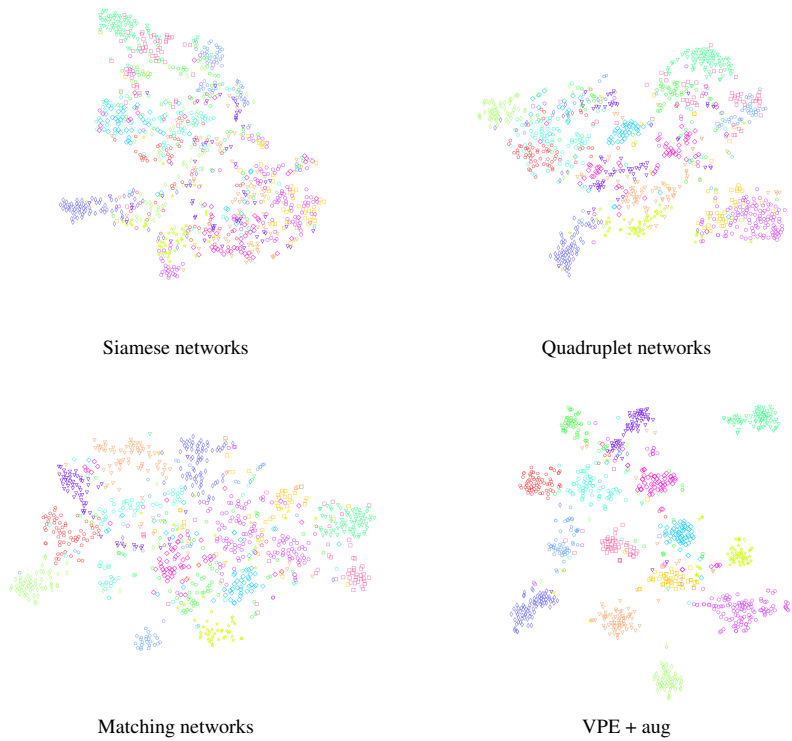

\vspace{-1mm}
\begin{center}
\setlength{\tabcolsep}{10pt}
\resizebox{0.69\linewidth}{!}{%
\large
\begin{tabular}{cc}
  
  \includegraphics[width=0.5\linewidth]{figure/belga2flickr_siamese.png} &
  \includegraphics[width=0.5\linewidth]{figure/belga2flickr_quadnet.png} \\
  Siamese networks & Quadruplet networks \\
  
  \includegraphics[width=0.5\linewidth]{figure/belga2flickr_matchingNet.png} &
  \includegraphics[width=0.5\linewidth]{figure/belga2flickr_VAE.png}\\
  Matching networks & VPE + aug \\
  
\end{tabular}
}
\end{center}
\vspace{-2mm}
   \caption{t-SNE visualization of features on embedding space. Features are randomly sampled from 15 different unseen classes under the Belga$\to$Flickr32 scenario for visualization.}
\vspace{-2.5mm}
\label{fig:tsne_logo}
\end{figure*}

\clearpage

\subsection{Image retrieval test}

We show more image retrieval results that could not be placed in the main text due to space constraints. The average images of the top $100$ images retrieved by querying unseen prototypes in each scenario are displayed. The columns from left to right are the average images retrieved using the Siamese networks \cite{koch2015siamese}, Quadruplet networks \cite{kim2018co}, Matching networks \cite{vinyals2016matching} and by the proposed method.

\begin{figure*}[h]
\vspace{-1mm}
\begin{center}
\setlength{\tabcolsep}{10pt}
\footnotesize
\resizebox{1.0\linewidth}{!}{%
\large
\begin{tabular}{ll}
  \multicolumn{2}{c}{GTSRB} \vspace{2mm} \\
  Prototype \hspace{0.8mm} Siamese \hspace{3.5mm} Quad \hspace{6.0mm} Match \hspace{6mm} VPE   &   Prototype \hspace{0.8mm} Siamese \hspace{3.5mm} Quad \hspace{6.0mm} Match \hspace{6mm} VPE
  \\
  \includegraphics[valign=t,width=0.5\linewidth]{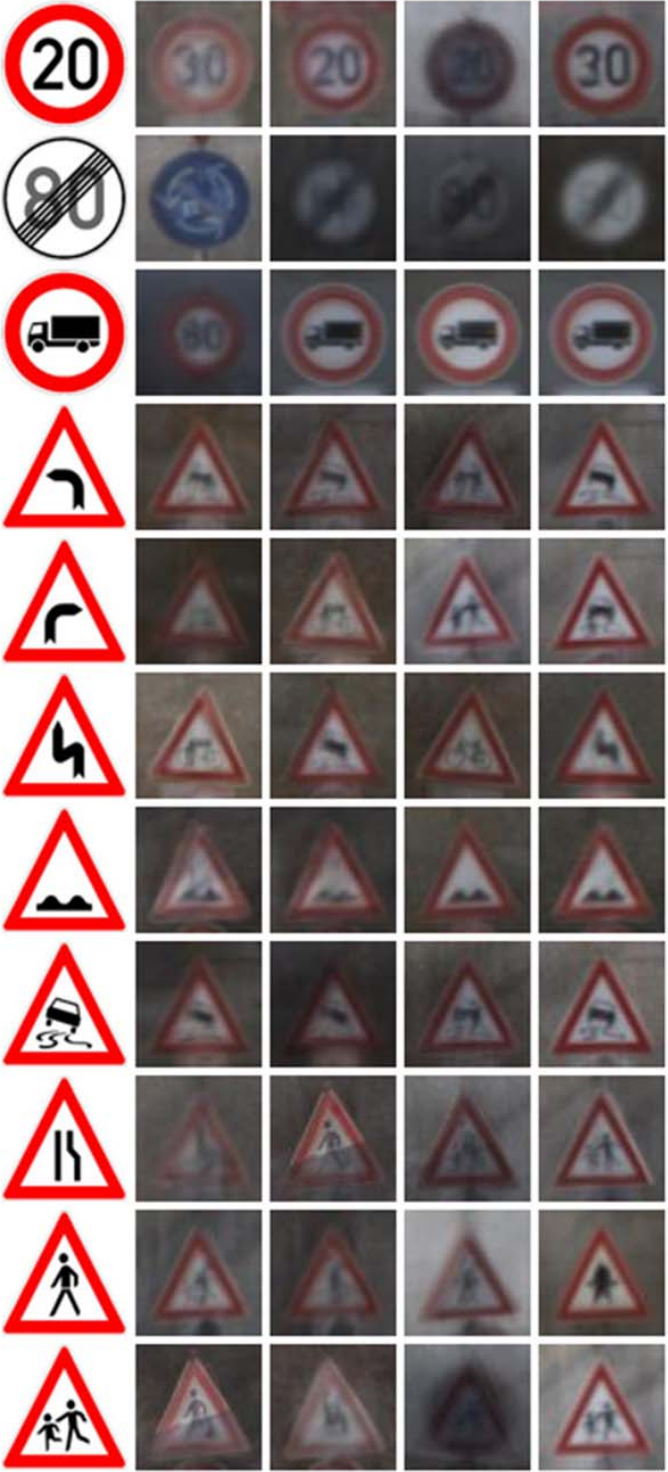} &
  \includegraphics[valign=t,width=0.5\linewidth]{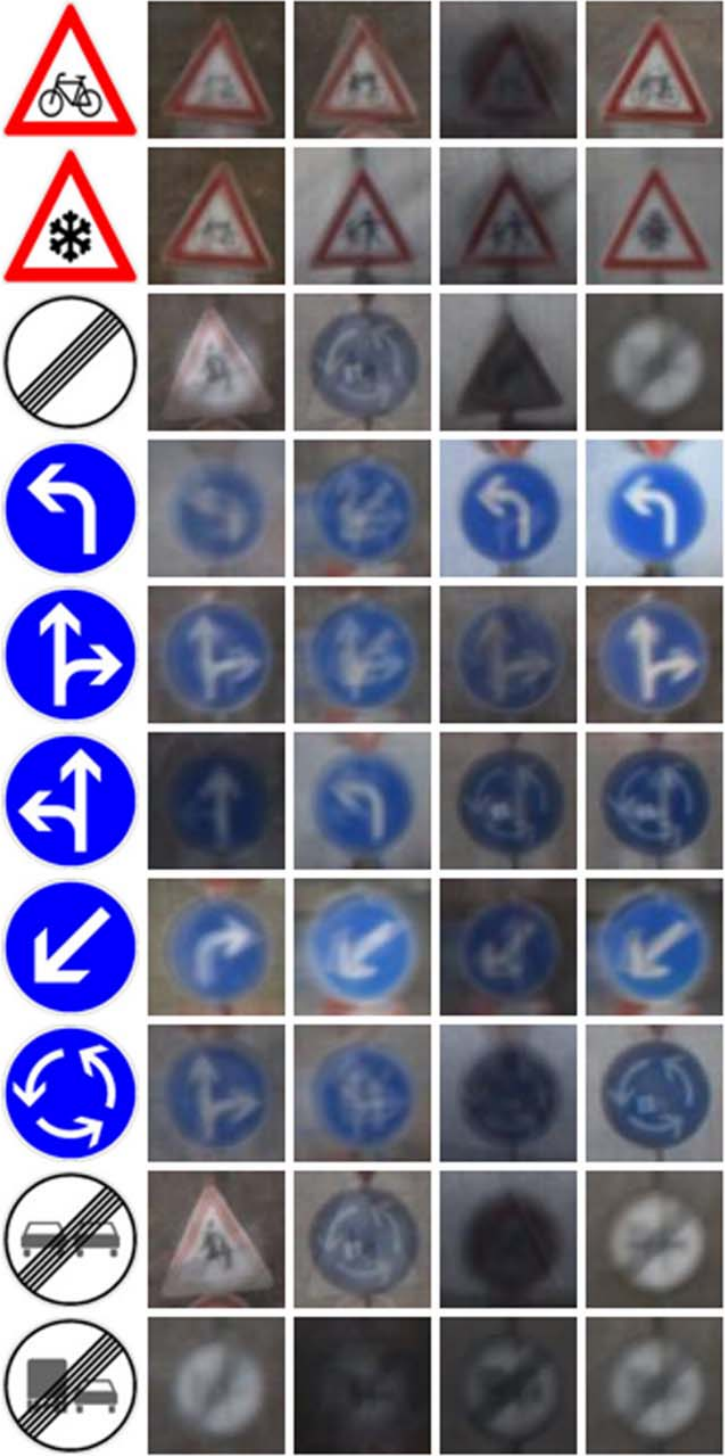} \\
  
\end{tabular}
}
\end{center}
\vspace{-2mm}
   \caption{Average images of top 100 retrieved images by querying unseen prototypes in the GTSRB scenario.}
\vspace{-2.5mm}
\label{fig:retrieve_gtsrb}
\end{figure*}

\begin{figure*}[h]
\vspace{-1mm}
\begin{center}
\setlength{\tabcolsep}{10pt}
\footnotesize
\resizebox{0.78\linewidth}{!}{%
\Large
\begin{tabular}{ll}

  \multicolumn{2}{c}{GTSRB$\to$TT100K} \vspace{2mm} \\
  \hspace{0.1mm} Proto \hspace{1.1mm} Siamese \hspace{1.9mm} Quad \hspace{3.7mm} Match \hspace{4mm} VPE   &   \hspace{0.1mm} Proto \hspace{1.1mm} Siamese \hspace{1.9mm} Quad \hspace{3.7mm} Match \hspace{4mm} VPE
  \\
  \includegraphics[width=0.5\linewidth]{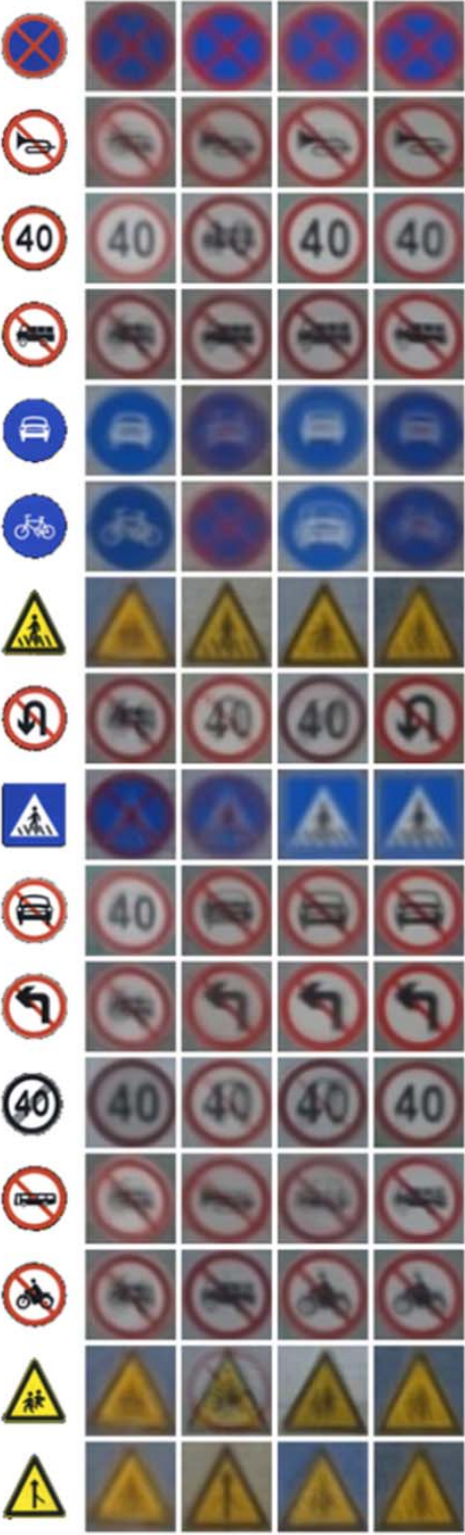} &
  \includegraphics[width=0.5\linewidth]{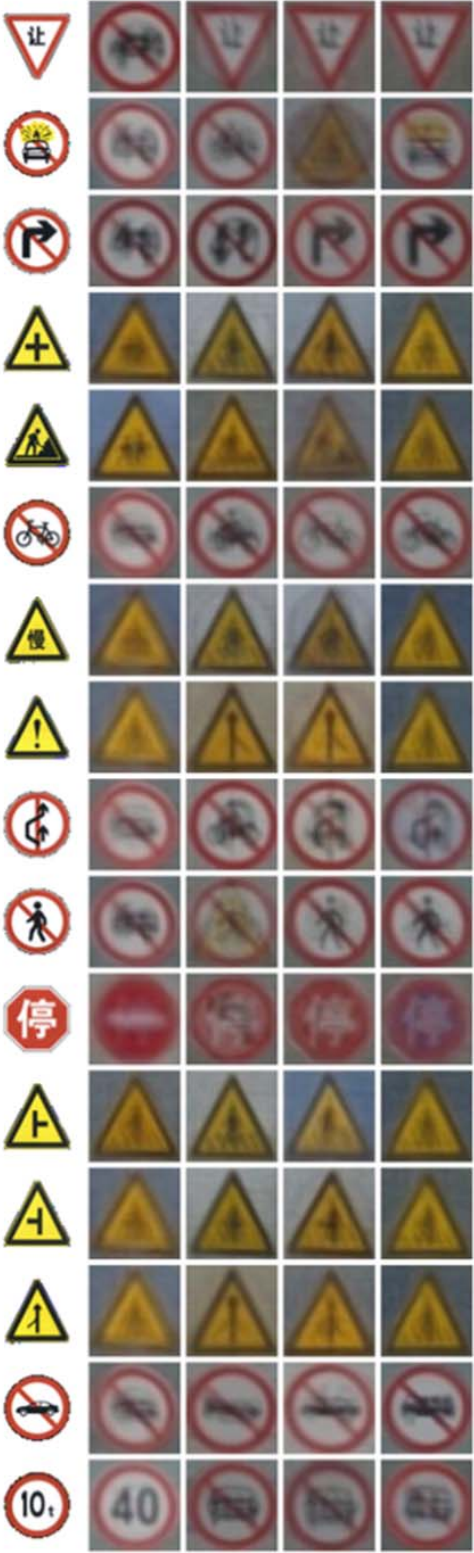} \\
  
\end{tabular}
}
\end{center}
\vspace{-2mm}
   \caption{Average images of top 100 retrieved images by querying unseen prototypes in the GTSRB$\to$TT100K scenario.}
\vspace{-2.5mm}
\label{fig:retrieve_gtsrb2TT100K}
\end{figure*}

\begin{figure*}[h]
\vspace{-1mm}
\begin{center}
\setlength{\tabcolsep}{10pt}
\footnotesize
\resizebox{0.9\linewidth}{!}{%
\Large
\begin{tabular}{ll}
  \multicolumn{2}{c}{Belga$\to$Flickr32} \vspace{2mm} \\
  \hspace{1.8mm} Proto \hspace{1.2mm} Siamese \hspace{1.1mm} Quad \hspace{3.0mm} Match \hspace{3.9mm} VPE & \hspace{1.8mm} Proto \hspace{1.2mm} Siamese \hspace{1.1mm} Quad \hspace{3.0mm} Match \hspace{3.9mm} VPE
  \\
  \includegraphics[width=0.5\linewidth]{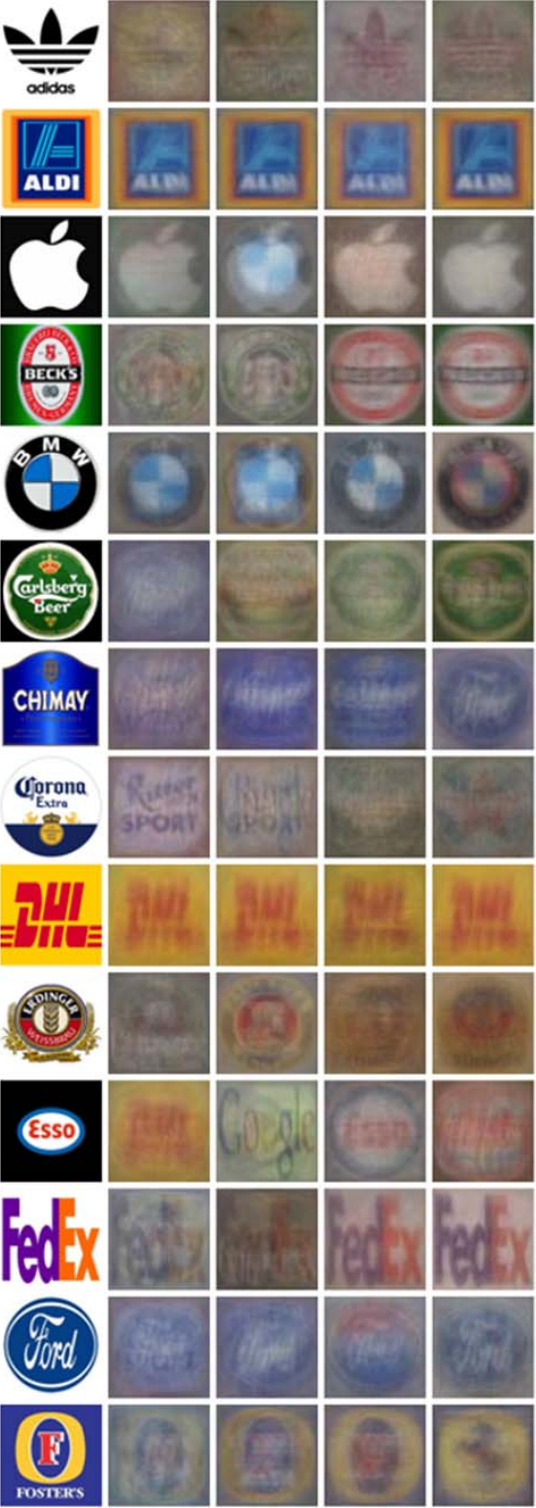} &
  \includegraphics[width=0.5\linewidth]{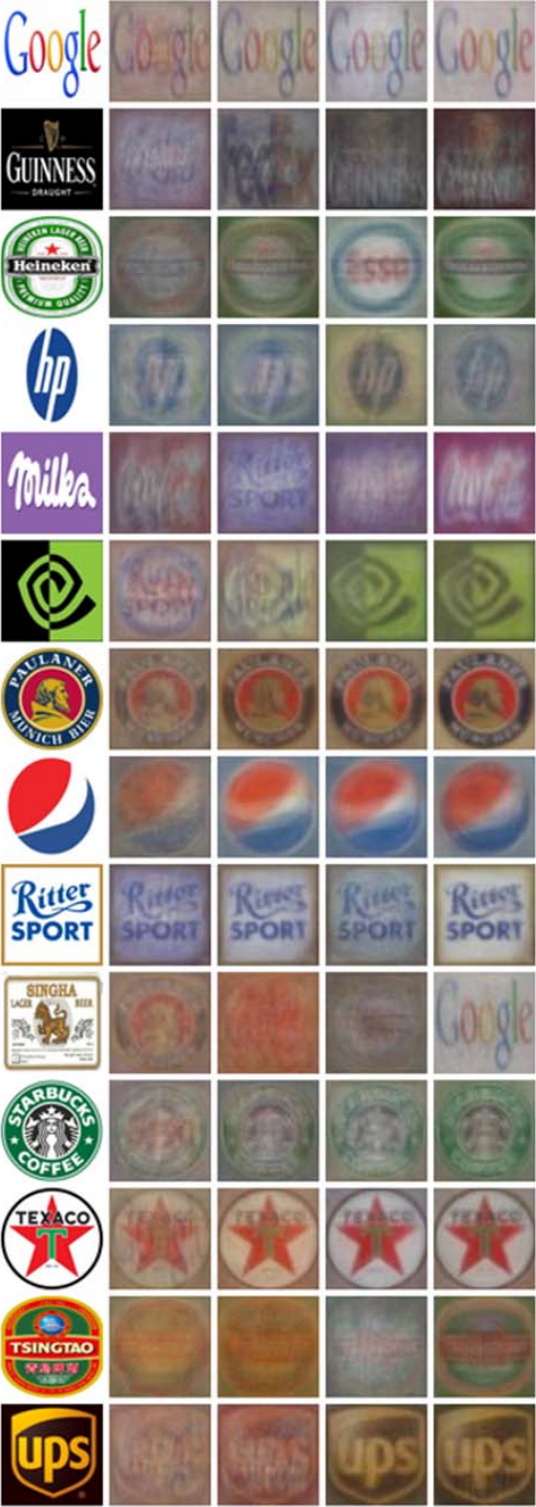} \\
  
\end{tabular}
}
\end{center}
\vspace{-2mm}
   \caption{Average images of top 100 retrieved images by querying unseen prototypes in the Belga$\to$Flickr32 scenario.}
\vspace{-2.5mm}
\label{fig:retrieve_belga2flickr}
\end{figure*}